\documentclass[11pt,letterpaper]{global_style}

\usepackage[utf8]{inputenc}
\usepackage[T1]{fontenc}
\usepackage{microtype}
\usepackage{graphicx}
\usepackage{booktabs}
\usepackage{float}
\usepackage{wrapfig}
\usepackage{adjustbox}
\usepackage{enumitem}
\usepackage{multirow}
\usepackage{subcaption}
\usepackage{nicefrac}
\usepackage{xspace}
\usepackage{url}
\usepackage{color}
\usepackage{xcolor}
\usepackage{makecell}
\usepackage{rotating}
\usepackage{fontawesome5}
\usepackage[font=small,labelfont=bf]{caption}
\usepackage[all]{hypcap}
\usepackage[comma,authoryear,compress]{natbib}
\usepackage{wrapfig} 
\definecolor{redxx}{HTML}{973901} 
\definecolor{borderbrown}{HTML}{631101}
\definecolor{Gray}{gray}{0.95}
\definecolor{darkgreen}{rgb}{0.0, 0.5, 0.0}

\usepackage{amsmath, amsfonts, amssymb}
\usepackage{mathtools}
\usepackage{amsthm}

\usepackage{algorithm}
\usepackage{algorithmicx}
\usepackage{algpseudocode}

\usepackage{hyperref}
\hypersetup{
  colorlinks = true,
  linkcolor = redxx,
  citecolor = redxx,
  urlcolor  = redxx
}

\newcommand{\keywordsblock}[1]{%
  {\color{redxx}\raisebox{-0.2ex}{\small\faTags}}\ \textbf{Keywords:} {\small #1}
}
\newcommand{\titleicon}{\color{redxx}\raisebox{0.3ex}{\small\faProjectDiagram}}

\usepackage{amssymb}
\usepackage{amsmath,amsfonts}
\usepackage{amsopn}
\usepackage{bm} 
\usepackage{multirow}
\usepackage{tabularx}

\newcommand{\ProbOpr}[1]{\mathbb{#1}}

\newcommand{\expect}[2]{%
\ifthenelse{\equal{#2}{}}{\ProbOpr{E}_{#1}}
{\ifthenelse{\equal{#1}{}}{\ProbOpr{E}\left[#2\right]}{\ProbOpr{E}_{#1}\left[#2\right]}}} 

\newcommand{\sD}{\mathcal{D}}

\newcommand{\sL}{\mathcal{L}}

\newcommand{\eat}[1]{}

\newcommand{\sM}{\mathcal{M}}

\newcommand{\sN}{\mathcal{N}}

\newcommand{\sP}{\mathcal{P}}
\newcommand{\sO}{\mathcal{O}}

\newcommand{\sR}{\mathcal{R}}

\title{\titleicon{} Context Matters: Learning Global Semantics via Object-Centric Representation}

\author[1]{Jike Zhong$^\dagger$}
\author[2]{Yuxiang Lai$^\dagger$}
\author[2]{Xiaofeng Yang}
\author[1]{Konstantinos Psounis}
\affil[1]{University of Southern California}
\affil[2]{Emory University}
\correspondingauthor{Jike Zhong (jikezhon@usc.edu)}

\begin{document}

\begin{abstract}

Recent advances in language modeling have witnessed the rise of highly desirable emergent capabilities, such as reasoning and in-context learning. However, vision models have yet to exhibit comparable progress in these areas. In this paper, we argue that this gap could stem from the lack of semantic and contextual guidance in current vision transformer (ViT) training schemes, and such a gap can be narrowed through the design of a semantic-grounded objective. Specifically, we notice that individual words in natural language are inherently semantic, and modeling directly on word tokens naturally learns a realistic distribution. In contrast, ViTs rely on spatial patchification, which inevitably lacks semantic information. To bridge this gap, we propose to directly model ``object" as the visual equivalence of ``word," pushing the model to learn the global context and semantics among visual elements. We investigate our hypotheses via masked image modeling (MIM), a framework where our approach can be readily tested by applying masks to visual objects rather than random patches. Considerable evidence from qualitative and quantitative evaluations reveals a key finding: object-level representation alone helps to learn a real-world distribution, whereas pixel-averaging shortcuts are often learned without it. Moreover, further evaluations with multimodal LLMs (MLLM) on visual question answering (VQA, GQA, ScienceQA) tasks demonstrate the strong reasoning and contextual understanding gained with this simple objective. We hope our study highlights effectiveness of object-level encoding and provides a plausible direction for developing stronger vision encoders and tokenizers. Code and model will be publicly released.

\vspace{1em}
\keywordsblock{Semantic Visual Tokenizer, Vision Reasoning, In-context Learning, Multimodal Reasoning}

\end{abstract}

\maketitle

\section{Introduction}

Recent studies have found that highly desirable capabilities such as reasoning and in-context learning can emerge naturally from the training process of large \emph{transformed}-based language models (LLMs) \citep{wei2022emergentabilitieslargelanguage, du2025understandingemergentabilitieslanguage, schaeffer2023emergentabilitieslargelanguage, wei2023chainofthoughtpromptingelicitsreasoning, kojima2023largelanguagemodelszeroshot, wang2024chainofthoughtreasoningprompting}, such as in Gemini \citep{geminiteam2024geminifamilyhighlycapable, geminiteam2024gemini15unlockingmultimodal}, BERT \citep{bert}, and GPT\citep{GPT-3, openai2024gpt4technicalreport}. These are surprising yet welcoming traits that enable promising downstream performance in many important areas---conversational AI, language agents, deep research, etc \citep{openai2024gpt4technicalreport, zhao2024surveylargelanguagemodels, Wang_2024, liu2024llmconversationalagentmemory, dam2024completesurveyllmbasedai}. 

In contrast, despite extensive work, vision transformers \citep{ViT} have yet to exhibit comparable emergent visual reasoning and in-context learning capabilities \citep{tong2024cambrian1fullyopenvisioncentric, bai2024sequential, MAE, bar2022visual}. Prior efforts have explored improving this through refined architectures \citep{liu2021swintransformerhierarchicalvision, wang2021pyramidvisiontransformerversatile, wu2021cvtintroducingconvolutionsvision, li2024visionlanguagemodelfinetuningsimple}, adjusted attention \citep{chu2021twinsrevisitingdesignspatial, yang2021focalselfattentionlocalglobalinteractions}, and multimodal training \citep{CLIP, fini2024multimodalautoregressivepretraininglarge, chen2025expandingperformanceboundariesopensource}, etc. In this work, we take a different approach and attempt to bridge this gap by identifying and minimizing the gap in the tokenization process between language and vision modeling. We start by re‑examining the inherent difference between natural language and vision; crucially, we identify a lack of explicit semantic guidance in ViT training. We then propose an object‑level objective within the masked image modeling (MIM) framework. We show that such a frustratingly simple semantic objective could fundamentally improve global contextual awareness, notably elevating visual reasoning and in-context learning.

To begin with, we summarize the key difference between language and visual modeling in \autoref{tab:vision language diff}. We notice that language modeling typically operates on discrete tokens (words) that inherently carry semantic meaning, allowing models to directly learn distributions over explicit semantic units and their contextual relationships \citep{GPT-3, bert, openai2023chatgpt}. In contrast, vision transformers (ViTs) tokenize images into spatially defined patches \citep{ViT} lacking inherent semantics, resulting in an initially continuous and semantically ambiguous distribution. While ViTs effectively learn semantics implicitly via attention and positional embeddings, the distribution they capture remains inherently less interpretable and less explicitly semantic compared to LLMs.

\begin{table*}[t]
    
    \centering
    \small
    \begin{tabular}{c|c|c|c|c}
    \toprule
         & Task Representation & Domain Gap & Information Density & \textbf{Representation Granularity} \\
    \midrule
        Language &  Unified & Small & Dense & \textbf{Semantic (\textcolor{redxx}{word})}\\
        Vision &  Task-Specific & Large & Sparse & \textbf{Structural (\textcolor{redxx}{patch})} \\
    \bottomrule
    \end{tabular}
    \caption{\textbf{Comparison of inherent properties of language and vision.}}
    \label{tab:vision language diff}
    \vskip -10pt
\end{table*}

\begin{wrapfigure}{r}{0.4\textwidth}
    \centering
    \vspace{-10pt}
    \includegraphics[width=\linewidth]{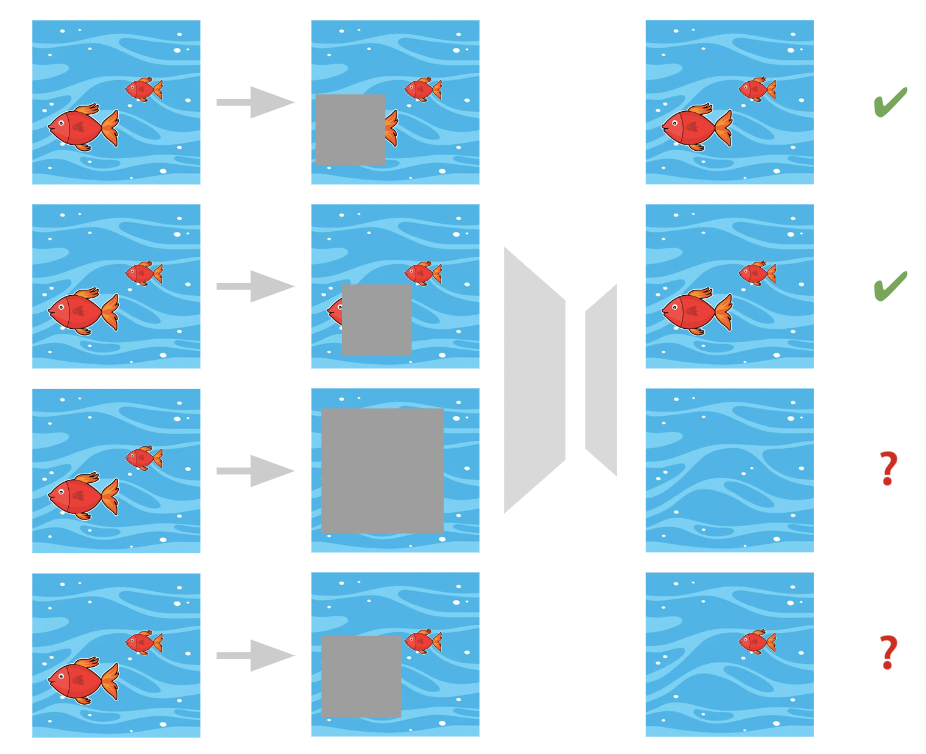}
    \caption{\textbf{By masking out random patches, current MIM setup encourages a shortcut learning where its generation is entirely based on surrounding pixels with little reference to global context.}}
    \label{fig:motivation}
    \vspace{-10pt}
\end{wrapfigure}

This observation presents a natural opportunity to bridge this gap through a more semantic tokenization process.  
However, it is not obvious how to design a semantic tokenizer. To efficiently investigate this issue, our approach is to leverage an existing framework for learning visual representation and apply an object-level objective. In selecting the framework, we first note that encoder-based frameworks such as CLIP \citep{CLIP} are hard to visualize without generation capability. On the other hand, pure decoder-style models, such as diffusion \citep{ho2020denoisingdiffusionprobabilisticmodels}, are difficult to integrate as the encoder into a multimodal system where downstream reasoning capability could be more broadly evaluated. We thus identify masked image modeling (MIM) as a suitable choice, as it provides both adaptabilities to visualization and downstream tasks.


Conventional MIMs \citep{pathak2016context} employ an encoder-decoder architecture, where an image with masked regions is encoded, and the missing content is reconstructed via decoding.
Recent efforts by \citet{MAE} extended the framework to ViTs by masking out random patches. However, such an approach bears exactly the aforementioned issue: as shown in \autoref{fig:motivation}, the model's prediction is ``nothing" 100\% of the time unless an object is partially visible, regardless of the context. This implies the model has not learned the actual object distribution. 
In contrast, instead of masking out random patches, we hypothesize ``object" as the visual equivalence of ``words" and incorporate a semantically grounded objective by masking entire objects explicitly. This effectively removes all potential object-based cues available and forces the model to learn global semantics by inferring the object using only the context. 

We evaluate our approach qualitatively and quantitatively. Visual prompting \citep{bar2022visual} for detection \citep{everingham2015pascal}, segmentation \citep{shaban2017one}, and scene completion demonstrates stronger contextual understanding, while downstream VQA tasks (VQA-V2 \citep{balanced_vqa_v2}, GQA \citep{hudson2019gqanewdatasetrealworld}, SQA \citep{lu2022learnexplainmultimodalreasoning}) using BLIP \citep{li2022blip, li2023blip2} and LLaVA \citep{liu2023visualinstructiontuning} confirm improved reasoning. Together, results suggest object-level representation enables learning realistic semantic distributions, whereas models without it tend to rely on shortcuts like “pixel-averaging.” \emph{To summarize, our contributions are as follows}:  

\begin{itemize}
    \item We identify the lack of explicit semantic guidance in tokenization as the key factor for the lack of reasoning and in-context learning capabilities in vision models.    
    \item We propose a simple yet effective object-level objective inspired by the success of language modeling and study it extensively through MIM. 
    \item We establish that vision models can learn semantics by learning the global image context and show that random masking encourages learning ``pixel-based shortcuts" rather than the true underlying distribution.
\end{itemize}

\begin{figure*}[t]
    \centering
    \minipage{0.9\linewidth}
    \centering
    \includegraphics[width=1\linewidth]{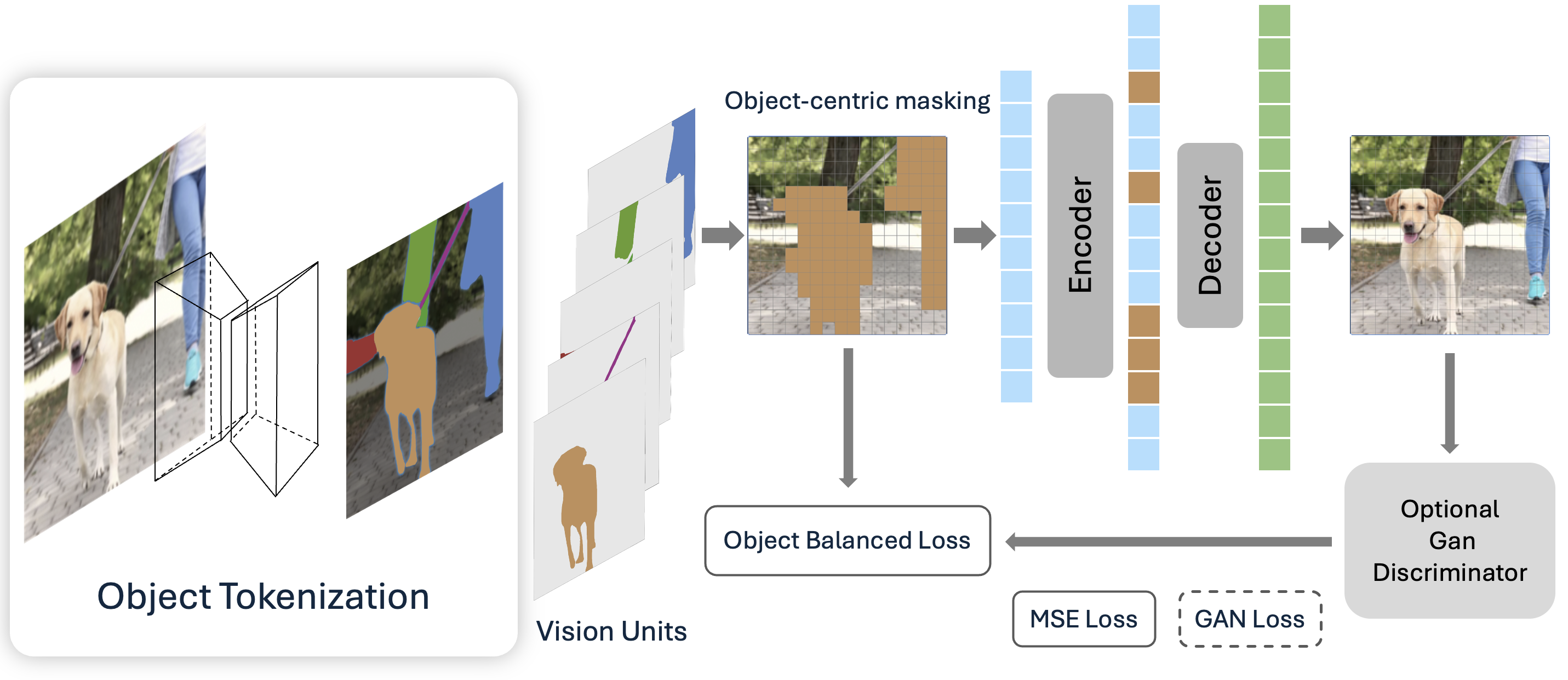} 
    \endminipage
    \caption{\textbf{Overall pipeline for Object-Centric MIM.} \textbf{We utilize a pre-trained segmentation model as an object tokenizer to segment the image into coarse object regions. The masked autoencoder is then trained using object-centric masking, and to further enhance the training of the object-centric encoder, we develop object-balanced loss.}} 
    \label{fig:pipeline}
    \vskip -10pt    
\end{figure*}

\section{Related Works}
\smallskip\noindent\textbf{Visual in-context learning.} This paradigm unifies diverse vision tasks, such as colorization, detection, and segmentation, into a single generative task \citep{bar2022visual, painter, wang2023seggpt, vicl1, zhang2023makes, li2023visual, foster2023flexible, zhang2023instruct}. It typically involves inpainting a grid-like prompt template with reference examples and query images as proposed in \citep{bar2022visual}. This naturally evaluates the model's capability of in-context learning as it requires the model to infer the correct answer based on the given contexts (example pair in this case). Visual in-context learning has become a popular alternative to fine-tuning for evaluating vision model's capacity \citep{lvm, sheng2023unified, sun2023exploring, wiedemer2025videomodelszeroshotlearners, lai2025patientspecificautoregressivemodelsorgan}, particularly in object-centric and contextual understanding. 

\smallskip\noindent\textbf{Masked image modeling (MIM).} MIM learns visual representations by reconstructing corrupted images with an encoder–decoder. Early work used CNNs \citep{Vincent2008, pathak2016context}, while images/{MAE} introduced transformers to recover masked patches. Later methods such as BEiT \citep{BeiT, mc-BEiT, PeCo, bar2022visual} predicted discrete tokens, and iBOT \citep{iBoT} and Siamese-MIM \citep{tao2022siamese} added contrastive objectives for global semantics. We adopt the transformer-based MIM framework to test our hypothesis, masking entire objects to enable “soft” semantic tokenization while retaining visualization and downstream adaptability. Our method is broadly compatible with existing MIM advances; the closest, \citet{semae}, masks parts of objects but we find it less effective for reasoning and in-context learning.

\section{Learning Global Context with Object-Level Representation}

\subsection{Masked Image Modeling}

\label{formulation}

We leverage the Masked Image Modelling (MIM) setup proposed by~\citet{he2022masked}, which is a transformer-based \citep{ViT} MIM framework, to implement and verify our approach.
The essential components of this framework includes an encoder and a decoder, where the encoder projects the unmasked input patches into latent representation, and the decoder decodes it along with the masked patches replaced with learnable mask tokens by by directly regressing on RGB pixel values.  

\smallskip\noindent\textbf{Setup.} Formally, an uncorrupted input image $x$ is first spatially tokenized into a sequence of $\sM$ total non-overlapping patches $\{x_{i}\}^{\sM}_{i=1}$ over all channels by tokenizer $q$. A random mask selection $m \in \{0,1\}^{\sM}$ is then applied to select $\sN=\sM r_{\mathrm{patch}}$ patches which will be masked out (removed from input), where $r_{\mathrm{patch}}$ is the predefined masking ratio and $m=1$ denotes a masked patch. The remaining visible patches sequence $\hat{x}_{\mathrm{patch}} =  \left\{ \hat{x}_i \middle| (1 - m_i)x_i\right\}_{i=1}^{\sM}$ thus forms the input for the encoder. For the decoder, the $\sN$ removed patches are first each replaced with a learnable mask token $e_{\left[mask\right]}$ and then placed back to the input sequence in location-aware manner.
An MSE loss is calculated over all corrupted $\sN$ patches only. 

\smallskip\noindent\textbf{Objective.} Let $\sD$ be the corpus. 
Let the end-to-end MIM model be parametrized by $\theta$. The objective is to maximize the following log-likelihood: 
\begin{equation}
\label{eq:maeobjective}
    \underset{\theta}{\max}\sum_{x\in\sD}{\mathbb{E}_{\sN} \left[ \sum_{i\in\sN}\log{\sP_{\theta}{(x_{i}|\hat{x}_{\mathrm{patch}})}} \right]}
\end{equation}
where $x_{i}$ denotes the missing patches to be reconstructed and $\hat{x}_{\mathrm{patch}}$ denotes the visible sequence of patches.  The overall objective is to train the autoencoder to reconstruct the missing patches using only the unmasked patches. Given that $x_i$ represents a patch, the minimal unit of reconstruction can be seen as done at patch level. Since the process of dividing an image into patches does not require any knowledge of the content, we can treat the tokenizer as a function parameterized by some constant $c$, usually patch size, such that:
\begin{equation}
    \label{eq:patch}
   x_{\mathrm{patch}} \coloneqq
 q{(x;c)}
\end{equation}

Note that here the tokenizer $q$ is simply a spatial divider, different from the canonical concept of tokenizer as found in \citet{BeiT, iBoT}.

\subsection{The Object-Centric Objective}
We now introduce our object-centric objective. The pipeline is illustrated in~\autoref{fig:pipeline}, with details outlined as follows.

\begin{figure*}[t]
    \centering
    \minipage{1\linewidth}
    \centering
    \includegraphics[width=1\linewidth]{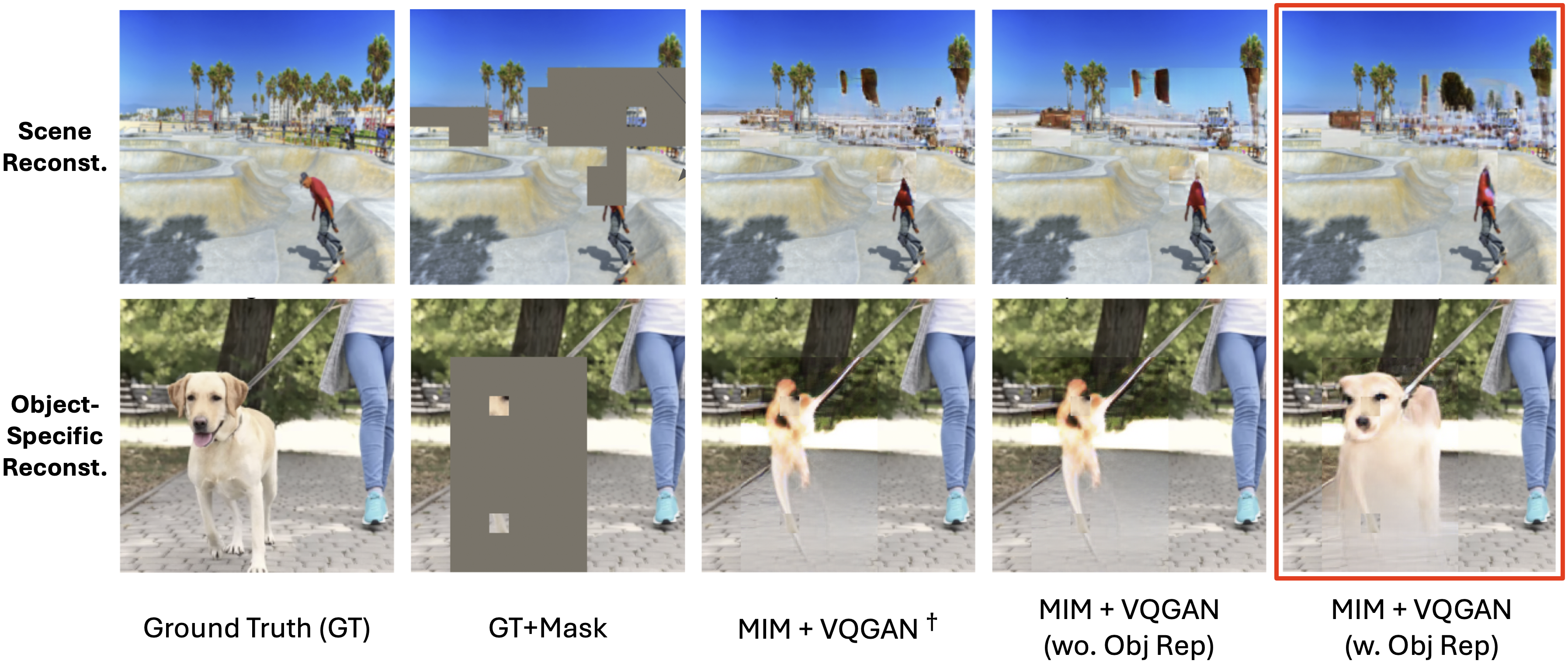} 
    \endminipage
    \caption{\textbf{ Reconstruction results of 1) scene reconstruction and 2) object-specific reconstruction.}} 
    \label{fig:main}
    \vskip -10pt
\end{figure*}

\label{appraoch:obj-mae}
\smallskip\noindent\textbf{Objective.} Let $\sR$ be the total number of objects $\{x_{j}\}^{\sR}_{j=1}$ in each image $x$ where $x_i$ represents an independent object. Let $\sO=\sR r_{\mathrm{obj}}$ be the total number of masked objects selected according to object-masking ratio $r_\mathrm{obj}$. We can rewrite the objective defined as \autoref{eq:maeobjective} with slight modification to reflect object-centric masking:
\begin{equation}
\label{eq:objobjective}
    \underset{\theta}{\max}\sum_{x\in\sD}{\mathbb{E}_{\sO} \left[ \sum_{j\in\sO}\log{\sP_{\theta}{(x_{j}|\hat{x}_{\mathrm\mathrm{obj}})}} \right]}
\end{equation}
where $x_j$ denotes the removed objects to be recovered and $\hat{x}_\mathrm{obj}$ denotes the image corrupted with some objects removed in their entirety. 

Although similar in form, the new objective differs significantly from the original \autoref{eq:maeobjective} in that it treats the object as the basic unit thus explicitly forcing the model to learn representation of objects based on the global context.

\smallskip\noindent\textbf{Mask generation and expansion.} 
\label{mask gen} To construct $\hat{x}_\mathrm{obj}$, we need to modify the data generation function \autoref{eq:patch} with new tokenizer: 
\begin{equation}
    x_{\mathrm{obj}} \coloneqq
 q'{(x; \phi)} 
\end{equation}
where $q'$ is the new tokenizer, represented by some network parametrized by $\phi$ that can generate coarse object masks dynamically for each image. Then, we can obtain $\hat{x}_{\mathrm{obj}}$ by masking out random objects. However, we empirically found this would easily lead to overfitting of object shape. We resolve this issue by expanding the coarse object mask to a square shape, essentially the bound box.

Since mask generation is separate from the model encoder, the learning of $q'(x;\phi)$ can be decoupled from training. As our main goal is to demonstrate the effectiveness of object-centric representation rather than learning a segmentation model, here we adopt the most efficient approach by applying the off-the-shelf pre-trained segmentation network SAM~\citep{kirillov2023segment}. Note that the usage of SAM is neither necessary nor required because only coarse masks are needed and we practically only need the bounding box; for example, we show that the same results could be achieved by using other fully unsupervised segmentation methods \citep{hahn2025scenecentricunsupervisedpanopticsegmentation}; please refer to the ablation study for detailed results.
Details on the specific mask generation procedure are provided in the Appendix.

\smallskip\noindent\textbf{Integration.} For each object $x_j$, first define the patch sequence $\Tilde{x}_j$ spatially representing the object. Since the size of the objects $s_j=|\Tilde{x}_j|$ differs, this problem can be effectively solved by fixing the total number of object pixels allowed per image. Since objects are entirely masked, the matrices $Q$, $K$, $V$ in the self-attention \citep{vaswani2017attention} module do not contain any info about the objects. Without the hints from the objects themselves, the decoder is forced to compute mask tokens solely based on global context, which effectively learns high-level features.

\subsection{Learning Context via Object-Centric Objective}
\label{two-stage}
\smallskip\noindent\textbf{Two-stage learning.}. Although \autoref{eq:objobjective} provides an intuitive objective, we empirically find that directly optimizing it is rather challenging due to the lack of prior knowledge on basic pixel reconstruction. Hence we propose a two-stage learning strategy, aiming to first learn easier low-level features and then learn the harder high-level features.

\smallskip\noindent\textbf{Optimization.} 
As the first stage is identical to plain MIM training, we can directly minimize the MSE loss as is. For the second stage, we add an object loss to account for varied object sizes (in terms of relative sizes in the image). 

Specifically, let $x_j \in \mathbb{R}^{3s_j\times{1}}$ be the ground truth pixel RGB value of object $j$ and $y_j\in \mathbb{R}^{3s_j\times{1}}$ be its predicted pixels values where $s\in\{s_j\}_{j=1}^{\sO}$ is a vector representing the sizes of the objects in terms of their pixel value. The first part of the loss \(\sL_{MIM}\) can then be written as: \begin{equation}
    \label{eq:maeloss}
    \sL_{MIM} = \frac{1}{\Omega(\mathbf{x}_\sO)}\sum_{j\in\sO}\sum_{k \in s_j} (y_j^k - x_j^k)^2
\end{equation}
where $\Omega(\cdot)$ denotes the total number of pixels of all corrupted objects in the image, $j$ denotes the object index, and $k$ denotes the pixel index. The second part is a balanced-object loss $\sL_\mathrm{obj}$ calculated based on a weight vector using a softmax function which maps unit-normalized $s$ to a relaxed probability vector inversely correlated with size, which can then be written as:
\begin{equation}
    \label{eq:objloss}
    \sL_\mathrm{obj} = \text{Softmax}(-\frac{s}{||s||})^{T}\cdot\left[\sum_{k \in s_j} (y_j^k - x_j^k)^2\right]_{j=1}^{\sO}
\end{equation}
Thus the second stage can be jointly optimized with a combination of \autoref{eq:maeloss} and \autoref{eq:objloss}:
\begin{equation}
    \label{eq:allloss}
    \sL_{OBJ-MIM} = \sL_{MIM} + \lambda_{1}\cdot\sL_\mathrm{obj}
\end{equation} 
where $\lambda_{1}$ is the scaling factor set to 0.4, which we empirically found to be the best.

\section{Experiments}
To demonstrate the effectiveness of our framework in learning semantics and contextual understanding, we carefully select three vision centric tasks for evaluation: 1) traditional vision tasks such as detection and segmentation via visual prompting and inpainting \citep{bar2022visual}, 2) scene-context reconstruction via inpainting, and 3) visual question answering \citep{balanced_vqa_v2, VQA} via multimodal visual instruction tuning (MLLM) \citep{tong2024cambrian1fullyopenvisioncentric}. These tasks are particularly suitable for evaluation in this case because they all require not only visual recognition but also spatial and compositional reasoning across the scene and context. Additionally, these tasks allows for both qualitative and quantitative measures, which we will discuss in detail in the remaining sections. Moreover, we also provide an additional toy study to further illustrate how our method can facilitate learning of visual contexts explicitly.

\begin{table*}[t]
\small
\centering
\setlength{\tabcolsep}{4pt}
\begin{tabular}
{l|cccc|cccc}

\toprule
\multirow{2}{*}{\textbf{Model}} & \multicolumn{4}{c}{\textbf{Foreground Segmentation $mIOU$ $\uparrow$}} & \multicolumn{4}{c}{\textbf{Single Object Detection $mIOU$ $\uparrow$}} \\
\cmidrule(r){2-9}
 & Split1 & Split2 & Split3 & Split4 & Split1 & Split2 & Split3 & Split4 \\
\midrule
BEiT$^*$ \citep{BeiT} & 5.38 & 3.94& 3.20 &3.29 &0.17& 0.02& 0.14& 0.16\\
MIM$^*$ \citep{MAE} & 17.42 & 25.7 & 18.64 & 16.53 & 5.49 & 4.98 & 5.24 & 5.84 \\
MIM (wo. Obj Rep) & 17.58 & 25.0 & 19.14 & 16.13 & 5.19 & 5.30 & 5.24 & 5.24 \\
MIM (w. Obj Rep) & 18.18 & 25.89 & 19.23 & 17.34 & 5.52 & 5.23 & 5.74 & 5.98 \\
MIM+VQGAN$^\dagger$ \citep{bar2022visual} & 27.83 & 30.64 & 26.15 & 24.00 & 24.20 & 25.2 & 25.35 & 25.12 \\
MIM+VQGAN (wo. Obj Rep) & 27.33 & 29.24 & 27.15 & 24.53 & 24.21 & 24.88 & 25.15 & 25.99 \\
MIM+VQGAN (w. Obj Rep) & \textbf{28.32} & \textbf{31.02} & \textbf{27.34} & \textbf{25.13} & \textbf{26.21} & \textbf{26.41} & \textbf{28.19} & \textbf{27.43} \\
\bottomrule
\end{tabular}
\caption{\textbf{Results for foreground segmentation and single object detection. ``$\dagger$" denotes direct evaluation or tuning with public checkpoints, $*$ denotes entries copied from \citet{bar2022visual}; \emph{notations apply to all subsquent experiments}. All other methods are trained using the same data.}}
\label{tab:seg_detect}
\vskip -10pt
\end{table*}

\subsection{Qualitative Evaluations}
\label{exp:qualitative eval}

\smallskip\noindent\textbf{Setup.} We evaluate our approach on two tasks: visual prompting \citep{bar2022visual}—feeding models a 4‑grid reference/query pair for copy, inpainting, colorization, and detection—and scene‑context reconstruction, which requires contextual and semantic understanding.

A naïve MAE‑style encoder–decoder on ImageNet \citep{Imagenet} would yield blurry generations and limited contextual learning due to its object‑centric nature. To address this, we follow \citet{bar2022visual}, using a VQGAN \citep{esser2021taming} to produce discrete visual tokens for sharper outputs, and adopt the scene‑centric SA1B \citep{kirillov2023segment} dataset to enrich context. All compared methods use the same additional data for fairness.

\smallskip\noindent\textbf{Implementation details.} 
\label{imple detail quali}
We largely follow the setup as in \citet{bar2022visual}. Specifically, we use ViT-Large-based \citep{ViT} models with 24 encoder blocks and 8 decoder blocks with a hidden embedding size of 1024 and 512. We resize image-mask pairs to $H\times W = 224 \times  224$ and adopt a patch size of p = 16. For VQGAN, we use the ImageNet \citep{Imagenet} pre-trained codebook as \citet{bar2022visual} with vocabulary size $|V| = 1024$. We train our models (initialized from publicly available checkpoints) on the pre-processed dataset with object-level representation for 50 epochs using 500K images. We use Adam \citep{AdamW} optimizer with cosine learning-rate schedule at an initial rate of $1e-5$. All experiments are conducted on a single Nvidia A100 GPU. Additional details can be found in the Appendix.

\begin{figure*}[t]
    \centering
    \minipage{1\linewidth}
    \centering
    \includegraphics[width=1\linewidth]{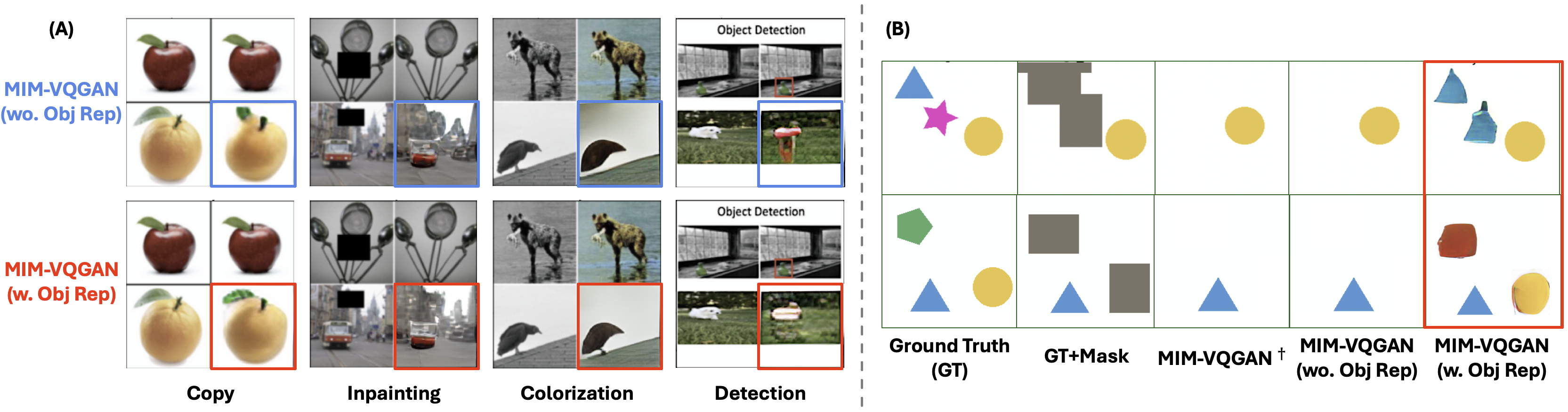} 
    \endminipage
    \caption{\textbf{(a) Visual in-context learning results on various vision tasks. (b) Reconstruction results on toy ``context" dataset.}} 
    \label{fig:in-context}
    \vskip -10pt
\end{figure*}

\smallskip\noindent\textbf{Analysis.} \label{sec: quali-analysis} \autoref{fig:in-context} shows the results for vision task and \autoref{fig:main} shows the results for scene-context composing/decomposing. Clearly, \autoref{fig:in-context} shows that our object-level objective approach facilitates better in-context learning capability compared to the original approach. For example, in the first column, our copy (bottom row) includes the leaf of the orange, which is absent in the upper row. The same pattern is also observed for the other three tasks. 

Moreover, \autoref{fig:main} demonstrates that our approach learns superior global semantics by leveraging the global contexts. For example, on the top row, the reconstruction of the original approach includes abrupt changes of sky color and unreasonable objects such as trees without trunks. In contrast, our method learns a natural filling of the missing area. On the bottom row, with minimum hint, our method correctly infers from the surroundings (the person and the leash) the existence of the dog, whereas the random masking objective only generates something totally unrecognizable. This reveals a major drawback of the original method: without proper object-level guidance, it merely learns a form of ``pixel-averaging," a shortcut instead of the true underlying distribution of visual elements. We further confirm this via a toy study in Toy Study Section. It is important to clarify that the benefits of our approach do not stem from the addition of training data, as results from columns 3 and 4 (original checkpoint vs. fine-tuned with new data via random masking) in \autoref{fig:main} are virtually identical.

\subsection{Quantitative Evaluations}
\label{exp:quantitative eval}
\smallskip\noindent\textbf{Setup.} We evaluate our method on two task groups: (1) traditional vision tasks via visual prompting, including foreground segmentation and single object detection \citep{bar2022visual}, and (2) visual question answering (VQA) \citep{VQA}. For vision tasks, we follow the qualitative evaluation setup and report mean IoU ($mIoU$) on Pascal-5i \citep{shaban2017one} and Pascal VOC 2012 \citep{everingham2015pascal}. For VQA, we pair our visual encoder with an MLLM tuned following BLIP \citep{li2022blip} and LLaVA \citep{liu2023visualinstructiontuning}, and evaluate on VQA-V2 \citep{balanced_vqa_v2}, GQA \citep{hudson2019gqanewdatasetrealworld}, and ScienceQA \citep{lu2022learnexplainmultimodalreasoning}, reporting average multiple-choice accuracy. Further benchmark details are in the Appendix.

\begin{table*}[t]
\centering
\small
\setlength{\tabcolsep}{3pt} 
\begin{minipage}[t]{0.48\textwidth}
\small
    \begin{tabular}{l|ccc}
        \toprule
        Model (w. LLaVA) & VQA (v2.0) & GQA & ScienceQA \\
        \midrule
        MIM$^\dagger$  & 53.02 & 36.24 & 39.04 \\
        SemMAE$^\dagger$  & 55.24 & 37.45 & 40.62 \\
        MIM (wo. Obj Rep) & 53.44 & 36.98 & 40.46 \\
        MIM (w. Obj Rep) & \textbf{56.89} & \textbf{40.00} & \textbf{42.98} \\
        \bottomrule
    \end{tabular}
    \caption{\textbf{Performance comparison across VQA (v2.0) \citep{balanced_vqa_v2}, GQA \citep{hudson2019gqanewdatasetrealworld}, and ScienceQA \citep{lu2022learnexplainmultimodalreasoning} with LLaVA \citep{liu2023visualinstructiontuning}. MIM \citep{MAE}, SemMAE \citep{li2022semmae}.}}
    \label{tab:performance_comparison}
\end{minipage}
\hfill
\begin{minipage}[t]{0.48\textwidth}
    \small
    \newcolumntype{C}[1]{>{\arraybackslash}p{#1}}
    \begin{tabular}{C{3cm}|C{0.9cm}C{0.9cm}C{0.9cm}|C{0.8cm}}    
    \toprule
        \multirow{3}{*}{Model (w. BLIP)} & \multicolumn{4}{c}{VQA (v2.0) Validation Acc (\%)} \\
    \cmidrule{2-5}
         & \multicolumn{3}{c|}{Fine-Grained Types} & \multirow{2}{*}{Overall}\\
         & Num. & Yes/No & Other   \\
    \midrule
        MIM$^\dagger$ & 36.22 & 71.00 & 41.32 & 51.80\\
        SemMAE$^\dagger$  & 36.28 & 70.97 & 41.37 & 51.85\\
        MIM (wo. Obj Rep) & 36.23 & 71.15 & 41.30 & 51.79\\
        MIM (w. Obj Rep) & \textbf{37.30} & \textbf{71.69} & \textbf{42.88}& \textbf{52.97} \\
    \bottomrule
    \end{tabular}
    \caption{\textbf{Fine-grained VQA (v2.0) \citep{balanced_vqa_v2} results with BLIP \citep{li2023blip2}.}}
    \label{tab:vqa}
\end{minipage}
\vskip -8pt
\end{table*}

\smallskip\noindent\textbf{Implementation details.}  
We follow the same setup as described in Implementation Details Section for encoder training.  
For visual instruction tuning \citep{liu2023visualinstructiontuning}, we first use LLaVA-V1.5-7B and follow the instruction tuning procedure from \citep{liu2023visualinstructiontuning}.  
Results are reported after two epochs of instruction tuning.  
Next, we adopt BLIP-V2 and fine-tune it using the same setup as~\citep{li2023blip2}. 

\smallskip\noindent\textbf{Analysis.} For pure vision tasks, our approach shows promising improvement across the board in foreground segmentation and single object detection, as shown in \autoref{tab:seg_detect}. Note that the addition of data does not help when random masking is employed and even hurts performance in some cases, which means the improvement seen in our approach stems directly from the object-level representation. 

\autoref{tab:vqa} shows the evaluation results with MLLM. Across the board, our approach demonstrates superior results. Notably, we observe up to 4\% improvement on VQA (v2) \citep{balanced_vqa_v2} and GQA \citep{hudson2019gqanewdatasetrealworld} with LLaVA \citep{liu2023visualinstructiontuning}. Many of these QA tasks require compositional understanding and reasoning beyond simple recognition. Improvements on these tasks further strengthen the observation that leveraging object-level representation to learn global context facilitates the learning of more semantic visual embeddings. We emphasize that our goal is not to achieve state-of-the-art (SOTA) results, but rather to explore whether improved objectives and tokenization can advance vision models.
We do not directly compare against CLIP~\citep{CLIP} as the encoder, since there is no straightforward way to integrate object-level representations directly into CLIP.
Notably, recent works have explored region-based alignment to improve CLIP’s localization capabilities~\citep{dong2023maskclipmaskedselfdistillationadvances, chen2025contrastivelocalizedlanguageimagepretraining, wan2024loccavisualpretraininglocationaware, naeem2023silcimprovingvisionlanguage}.
Nevertheless, these results do not contradict our findings.

\subsection{Toy Study}
\label{sec: toy study}
\smallskip\noindent\textbf{Setup.} Building on the previous qualitative and quantitative evaluations, we conduct a toy study to explicitly examine contextual learning. We create a “shape” dataset with five shapes, where the yellow circle and blue triangle always co-occur and other shapes serve as distractors. The model is trained to infer the missing object in the context pair when only one is visible. All shapes appear with equal frequency. We generate 200 training images, train for 100 epochs, and report results in \autoref{tab:toy} and \autoref{fig:in-context}.

\begin{table}[H]
\centering
\small
\setlength{\tabcolsep}{4pt} 
\begin{tabular}{l|c}
\toprule
    Model & Context Recovery Rate (\%)  \\
\midrule
    MIM+VQGAN$^\dagger$ \citep{bar2022visual} & 0.00 \\
    MIM+VQGAN (wo. Obj Rep) & 0.00\\
    MIM+VQGAN (w. Obj Rep) & \textbf{93.25} \\
\bottomrule
\end{tabular}
\caption{\textbf{Contextual pair recovery results. Our model recovers exclusive contexts 100\% while other models simply fail to recover any context, signalling that global semantics has been learned.}}
\label{tab:toy}
\vspace{-10pt}
\end{table}


\smallskip\noindent\textbf{Analysis.} 
\label{sec: toy-analysis} The last column in \autoref{fig:in-context} shows that our object-level objective approach correctly learns the contextual relationship as it is able to recover the ``blue triangle" given merely the ``yellow circle", or vice versa, 93\% of the times (\autoref{tab:toy}), compared to 0\% without object-level representation. Moreover, we emphasize two key observations here. First, besides the context pair, the other object being generated could be ``anything" or even ``nothing" (which is valid) since the underlying distribution only dictates the contextual pair but does not constrain the remaining objects. Second, the model trained with random masking objectively would generate ``nothing" 100\% of the time (\autoref{tab:toy}), completely neglecting the contextual relationship among the objects. This confirms that the model only learns a shortcut for generation through ``pixel-averaging" (similar to finding in \autoref{sec: quali-analysis}, contrary to the true underlying distribution among the objects that our approach is able to learn.
   

\section{Ablation Study \& Discussion}
\label{discussion}
In this section, we address common concerns and further discuss our key findings.

 \smallskip\noindent\textbf{Does the gain come primarily from object-level tokenization or SAM?}
Since we used SAM to obtain the object masks, it is reasonable to ask whether the improvement stems from the use of object-level tokenization during training or from the high precision of SAM’s object masks.
We emphasize that SAM is neither necessary nor required, as our method only relies on coarse masks that roughly cover the objects, rather than fine-grained, pixel-level annotations.
For full rigor and transparency, we dissect this factor by ablating the mask source: we replace SAM with a fully unsupervised segmentation network \citep{hahn2025scenecentricunsupervisedpanopticsegmentation}. We run inference on the entire training set using this unsupervised method to obtain object masks, then follow the same experimental setup.
The results in \autoref{tab:ablation} show that our method achieves comparable performance using unsupervised masks, confirming that the performance gain arises from object-level tokenization during training rather than from the precision of the masks themselves.

\begin{table}[H]
\centering
\small
\setlength{\tabcolsep}{4pt} 
\begin{tabular}{l|ccc}
\toprule
Model (w. LLaVA) & VQA (v2.0) & GQA & ScienceQA \\
\midrule
MIM (mask w/ SAM) & 56.89 & \textbf{40.00} & \textbf{42.98} \\
MIM (mask w/o SAM) & \textbf{57.66} & 39.12 & 42.56 \\
\bottomrule
\end{tabular}
\caption{\textbf{Ablation study: object mask obtained using/without using SAM\citep{kirillov2023segment}. Results demonstrate that improvement is NOT tied to SAM.}}
\label{tab:ablation}
\vspace{-10pt}
\end{table}

\begin{figure}[t]
    \centering
    \vspace{-10pt}
    \includegraphics[width=0.4\linewidth]{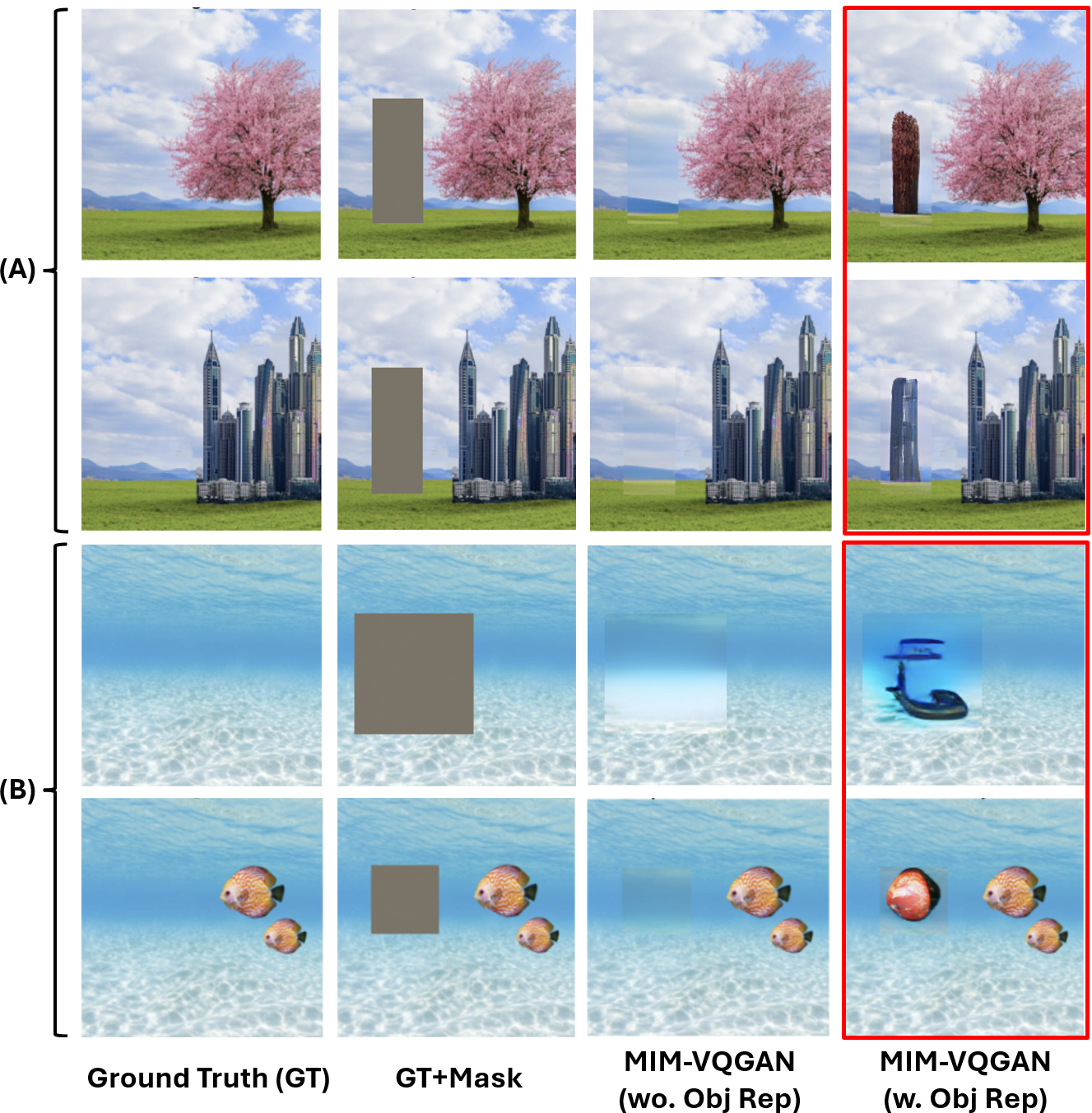}
\caption{\textbf{(A) Object Definition and Context: The representations of objects are learned based on ``Color'' and ``Shape.'' 
(B) Minimal Context Reconstruction: Reconstruction performed with minimal context, both with and without object reference.}}
\label{fig:color_shape_minimal}
    \vspace{-10pt}
\end{figure}

\smallskip\noindent\textbf{Finding 1: Semantics can be learned explicitly in vision models by learning global context.}  Evidence from \autoref{fig:main} and \autoref{fig:in-context} (b) show that by learning with object-level representation, the vision model will be able to learn contextual relationships. We provide more visualization in \autoref{fig:color_shape_minimal}. The top blocks essentially show the model can infer from ``color" and ``shape" key factors in how humans perceive objects \citep{reppa2020relative}. The bottom block shows that the model can generate objects based on the true distribution even with minimal context. Note that with different seeds, the results could be drastically different: while objects could be generated based on the context, on some occasions, no object could be generated. This is valid because both cases exist in true distribution. 

\smallskip\noindent\textbf{Finding 2: Tasks that explicitly require contextual understanding and reasoning benefit the most from object-level representation, while tasks that do not rely on context remain unaffected} While our approach enhances vision reasoning, there is a need to confirm that it does not degrade recognition. To validate, we adopt the encoder from Quantitative Evaluation Section and provide linear-probing (LP) and fine-tuning (FT) results on ImageNet-1K \citep{Imagenet}. The results are shown in \autoref{tab:cifar100}, and our method shows minor improvement in both settings. Notably, we observe that training on the additional data with the original objective (random masking) suffers significant degradation (-16.71\% and -15.94\%) compared to our approach and the original checkpoint, which is consistent with our findings in Quantitative Evaluation Section.
On the one hand, this suggests that additional data alone does not lead to performance improvement; on the other hand, it demonstrates strong generalization ability of our approach, which serves as valuable byproduct for vision models.

\begin{table}[H]
\centering
\small
\setlength{\tabcolsep}{8pt} 
\renewcommand{\arraystretch}{1.2} 
\begin{tabular}{c|cc}
\toprule
\multirow{2}{*}{\textbf{Model}} & \multicolumn{2}{c}{\textbf{ImageNet-1K Top-1 Acc (\%)}} \\
\cmidrule{2-3}
 & \textbf{FT} & \textbf{LP} \\
\midrule
$\text{MIM}^\dagger$ \citep{he2022masked} & 83.66 & 70.80 \\
$\text{SemMAE}^\dagger$ \citep{li2022semmae} & 83.73 & 71.25 \\
$\text{MIM (wo. Obj Rep)}$ & 67.72\ \textcolor{red}{\scriptsize{$\downarrow$15.94}} & 58.75\ \textcolor{red}{\scriptsize{$\downarrow$12.05}} \\
$\text{MIM (w. Obj Rep)}$ & \textbf{84.43\ \textcolor{darkgreen}{\scriptsize{$\uparrow$0.77}}} & \textbf{71.91\ \textcolor{darkgreen}{\scriptsize{$\uparrow$1.11}}} \\
\bottomrule
\end{tabular}
\caption{\textbf{Linear probing (LP) and fine-tuning (FT) results on ImageNet-1K. ``$\dagger$'' denotes direct LP/FT with public checkpoints \citep{Krizhevsky2009LearningML}}.}
\label{tab:cifar100}
\vskip -10pt
\end{table}

\smallskip\noindent\textbf{Finding 3. Random masking encourages learning a ``pixel-based shortcut" rather than the true distribution.}  
As shown in \autoref{fig:main}, \autoref{fig:in-context}, and \autoref{fig:color_shape_minimal}, while our approach learns to generate based on a meaningful underlying distribution, random masking results in no object being generated unless it is partially visible.  
This suggests the model learns a ``pixel-based shortcut" akin to interpolation rather than capturing true relationships and semantics.


\section{Conclusion} In this work, we provide a study into whether object-level representation could facilitate the learning of global semantics and contexts, thus enhancing vision models' contextual reasoning and understanding capability. Through our qualitative evaluation via visual prompting and quantitative evaluation via MLLM, we demonstrate that this objective is indeed useful. We hope our study not only provides insight into enhancing visual reasoning but also how we can improve the generalizability and scalability of vision models in general via better tokenization. 

\clearpage
\section{Acknowledgment}
This work is supported by an award from the USC and Amazon Center on Secure \& Trusted Machine Learning. We also thank Yutong Bai and Alan Yuille for their helpful discussions.
\bibliography{main}

\begin{thebibliography}{84}
\providecommand{\natexlab}[1]{#1}
\providecommand{\url}[1]{\texttt{#1}}
\expandafter\ifx\csname urlstyle\endcsname\relax
  \providecommand{\doi}[1]{doi: #1}\else
  \providecommand{\doi}{doi: \begingroup \urlstyle{rm}\Url}\fi

\bibitem[Antol et~al.(2015)Antol, Agrawal, Lu, Mitchell, Batra, Zitnick, and Parikh]{VQA}
Stanislaw Antol, Aishwarya Agrawal, Jiasen Lu, Margaret Mitchell, Dhruv Batra, C.~Lawrence Zitnick, and Devi Parikh.
\newblock Vqa: Visual question answering.
\newblock In \emph{International Conference on Computer Vision (ICCV)}, 2015.

\bibitem[Bai et~al.(2023)Bai, Geng, Mangalam, Bar, Yuille, Darrell, Malik, and Efros]{lvm}
Yutong Bai, Xinyang Geng, Karttikeya Mangalam, Amir Bar, Alan Yuille, Trevor Darrell, Jitendra Malik, and Alexei~A Efros.
\newblock Sequential modeling enables scalable learning for large vision models, 2023.

\bibitem[Bai et~al.(2024)Bai, Geng, Mangalam, Bar, Yuille, Darrell, Malik, and Efros]{bai2024sequential}
Yutong Bai, Xinyang Geng, Karttikeya Mangalam, Amir Bar, Alan~L Yuille, Trevor Darrell, Jitendra Malik, and Alexei~A Efros.
\newblock Sequential modeling enables scalable learning for large vision models.
\newblock In \emph{Proceedings of the IEEE/CVF Conference on Computer Vision and Pattern Recognition}, pages 22861--22872, 2024.

\bibitem[Bao et~al.(2022)Bao, Dong, Piao, and Wei]{BeiT}
Hangbo Bao, Li~Dong, Songhao Piao, and Furu Wei.
\newblock {BE}it: {BERT} pre-training of image transformers.
\newblock In \emph{ICLR}, 2022.

\bibitem[Bar et~al.(2022)Bar, Gandelsman, Darrell, Globerson, and Efros]{bar2022visual}
Amir Bar, Yossi Gandelsman, Trevor Darrell, Amir Globerson, and Alexei Efros.
\newblock Visual prompting via image inpainting.
\newblock \emph{Advances in Neural Information Processing Systems}, 35:\penalty0 25005--25017, 2022.

\bibitem[Bartnik and Groen(2023)]{BartnikGroen2023}
Clemens~G. Bartnik and Iris I.~A. Groen.
\newblock \emph{Visual perception in the human brain: How the brain perceives and understands real‐world scenes}.
\newblock Oxford University Press, 2023.
\newblock \doi{10.1093/acrefore/9780190264086.013.437}.

\bibitem[Bonner and Epstein(2021)]{BonnerEpstein2021}
Michael~F. Bonner and Russell~A. Epstein.
\newblock Object representations in the human brain reflect the co‐occurrence statistics of vision and language.
\newblock \emph{Nature Communications}, 12\penalty0 (1):\penalty0 4081, 2021.
\newblock \doi{10.1038/s41467-021-24368-2}.

\bibitem[Brown et~al.(2020)Brown, Mann, Ryder, Subbiah, Kaplan, Dhariwal, Neelakantan, Shyam, Sastry, Askell, et~al.]{GPT-3}
Tom Brown, Benjamin Mann, Nick Ryder, Melanie Subbiah, Jared~D Kaplan, Prafulla Dhariwal, Arvind Neelakantan, Pranav Shyam, Girish Sastry, Amanda Askell, et~al.
\newblock Language models are few-shot learners.
\newblock \emph{NeurIPS}, 33:\penalty0 1877--1901, 2020.

\bibitem[Chen et~al.(2025{\natexlab{a}})Chen, Lai, Zhang, Wang, Eichner, You, Cao, Zhang, Yang, and Gan]{chen2025contrastivelocalizedlanguageimagepretraining}
Hong-You Chen, Zhengfeng Lai, Haotian Zhang, Xinze Wang, Marcin Eichner, Keen You, Meng Cao, Bowen Zhang, Yinfei Yang, and Zhe Gan.
\newblock Contrastive localized language-image pre-training, 2025{\natexlab{a}}.
\newblock URL \url{https://arxiv.org/abs/2410.02746}.

\bibitem[Chen et~al.(2023)Chen, Song, Yeo, Liu, Fu, and Shuai]{vicl1}
Yi-Syuan Chen, Yun-Zhu Song, Cheng~Yu Yeo, Bei Liu, Jianlong Fu, and Hong-Han Shuai.
\newblock Sinc: Self-supervised in-context learning for vision-language tasks, 2023.

\bibitem[Chen et~al.(2025{\natexlab{b}})Chen, Wang, Cao, Liu, Gao, Cui, Zhu, Ye, Tian, Liu, Gu, Wang, Li, Ren, Chen, Luo, Wang, Jiang, Wang, He, Shi, Zhang, Lv, Wang, Shao, Chu, Tu, He, Wu, Deng, Ge, Chen, Zhang, Wang, Dou, Lu, Zhu, Lu, Lin, Qiao, Dai, and Wang]{chen2025expandingperformanceboundariesopensource}
Zhe Chen, Weiyun Wang, Yue Cao, Yangzhou Liu, Zhangwei Gao, Erfei Cui, Jinguo Zhu, Shenglong Ye, Hao Tian, Zhaoyang Liu, Lixin Gu, Xuehui Wang, Qingyun Li, Yimin Ren, Zixuan Chen, Jiapeng Luo, Jiahao Wang, Tan Jiang, Bo~Wang, Conghui He, Botian Shi, Xingcheng Zhang, Han Lv, Yi~Wang, Wenqi Shao, Pei Chu, Zhongying Tu, Tong He, Zhiyong Wu, Huipeng Deng, Jiaye Ge, Kai Chen, Kaipeng Zhang, Limin Wang, Min Dou, Lewei Lu, Xizhou Zhu, Tong Lu, Dahua Lin, Yu~Qiao, Jifeng Dai, and Wenhai Wang.
\newblock Expanding performance boundaries of open-source multimodal models with model, data, and test-time scaling, 2025{\natexlab{b}}.
\newblock URL \url{https://arxiv.org/abs/2412.05271}.

\bibitem[Chu et~al.(2021)Chu, Tian, Wang, Zhang, Ren, Wei, Xia, and Shen]{chu2021twinsrevisitingdesignspatial}
Xiangxiang Chu, Zhi Tian, Yuqing Wang, Bo~Zhang, Haibing Ren, Xiaolin Wei, Huaxia Xia, and Chunhua Shen.
\newblock Twins: Revisiting the design of spatial attention in vision transformers, 2021.
\newblock URL \url{https://arxiv.org/abs/2104.13840}.

\bibitem[Dam et~al.(2024)Dam, Hong, Qiao, and Zhang]{dam2024completesurveyllmbasedai}
Sumit~Kumar Dam, Choong~Seon Hong, Yu~Qiao, and Chaoning Zhang.
\newblock A complete survey on llm-based ai chatbots, 2024.
\newblock URL \url{https://arxiv.org/abs/2406.16937}.

\bibitem[Deng et~al.(2009)Deng, Dong, Socher, Li, Li, and Fei-Fei]{Imagenet}
Jia Deng, Wei Dong, Richard Socher, Li-Jia Li, Kai Li, and Li~Fei-Fei.
\newblock Imagenet: A large-scale hierarchical image database.
\newblock In \emph{CVPR}, pages 248--255, 2009.

\bibitem[Devlin et~al.(2018)Devlin, Chang, Lee, and Toutanova]{bert}
Jacob Devlin, Ming-Wei Chang, Kenton Lee, and Kristina Toutanova.
\newblock Bert: Pre-training of deep bidirectional transformers for language understanding.
\newblock \emph{arXiv preprint arXiv:1810.04805}, 2018.

\bibitem[Dong et~al.(2021)Dong, Bao, Zhang, Chen, Zhang, Yuan, Chen, Wen, and Yu]{PeCo}
Xiaoyi Dong, Jianmin Bao, Ting Zhang, Dongdong Chen, Weiming Zhang, Lu~Yuan, Dong Chen, Fang Wen, and Nenghai Yu.
\newblock Peco: Perceptual codebook for bert pre-training of vision transformers.
\newblock \emph{arXiv preprint arXiv:2111.12710}, 2021.

\bibitem[Dong et~al.(2023)Dong, Bao, Zheng, Zhang, Chen, Yang, Zeng, Zhang, Yuan, Chen, Wen, and Yu]{dong2023maskclipmaskedselfdistillationadvances}
Xiaoyi Dong, Jianmin Bao, Yinglin Zheng, Ting Zhang, Dongdong Chen, Hao Yang, Ming Zeng, Weiming Zhang, Lu~Yuan, Dong Chen, Fang Wen, and Nenghai Yu.
\newblock Maskclip: Masked self-distillation advances contrastive language-image pretraining, 2023.
\newblock URL \url{https://arxiv.org/abs/2208.12262}.

\bibitem[Dosovitskiy et~al.(2021)Dosovitskiy, Beyer, Kolesnikov, Weissenborn, Zhai, Unterthiner, Dehghani, Minderer, Heigold, Gelly, Uszkoreit, and Houlsby]{ViT}
Alexey Dosovitskiy, Lucas Beyer, Alexander Kolesnikov, Dirk Weissenborn, Xiaohua Zhai, Thomas Unterthiner, Mostafa Dehghani, Matthias Minderer, Georg Heigold, Sylvain Gelly, Jakob Uszkoreit, and Neil Houlsby.
\newblock An image is worth 16x16 words: Transformers for image recognition at scale.
\newblock In \emph{ICLR}, 2021.

\bibitem[Du et~al.(2025)Du, Zeng, Dong, and Tang]{du2025understandingemergentabilitieslanguage}
Zhengxiao Du, Aohan Zeng, Yuxiao Dong, and Jie Tang.
\newblock Understanding emergent abilities of language models from the loss perspective, 2025.
\newblock URL \url{https://arxiv.org/abs/2403.15796}.

\bibitem[Esser et~al.(2021)Esser, Rombach, and Ommer]{esser2021taming}
Patrick Esser, Robin Rombach, and Bj{\"o}rn Ommer.
\newblock Taming transformers for high-resolution image synthesis.
\newblock \emph{Proceedings of the IEEE/CVF Conference on Computer Vision and Pattern Recognition (CVPR)}, pages 12873--12883, 2021.
\newblock \doi{10.1109/CVPR46437.2021.01268}.
\newblock URL \url{https://arxiv.org/abs/2012.09841}.

\bibitem[Everingham et~al.(2015)Everingham, Eslami, Van~Gool, Williams, Winn, and Zisserman]{everingham2015pascal}
Mark Everingham, SM~Ali Eslami, Luc Van~Gool, Christopher~KI Williams, John Winn, and Andrew Zisserman.
\newblock The pascal visual object classes challenge: A retrospective.
\newblock \emph{International Journal of Computer Vision}, 111\penalty0 (1):\penalty0 98--136, 2015.

\bibitem[Fei et~al.(2023)Fei, Fan, Zhu, Huang, Wei, and Wei]{fei2023masked}
Zhengcong Fei, Mingyuan Fan, Li~Zhu, Junshi Huang, Xiaoming Wei, and Xiaolin Wei.
\newblock Masked auto-encoders meet generative adversarial networks and beyond.
\newblock In \emph{Proceedings of the IEEE/CVF Conference on Computer Vision and Pattern Recognition}, pages 24449--24459, 2023.

\bibitem[Fini et~al.(2024)Fini, Shukor, Li, Dufter, Klein, Haldimann, Aitharaju, da~Costa, Béthune, Gan, Toshev, Eichner, Nabi, Yang, Susskind, and El-Nouby]{fini2024multimodalautoregressivepretraininglarge}
Enrico Fini, Mustafa Shukor, Xiujun Li, Philipp Dufter, Michal Klein, David Haldimann, Sai Aitharaju, Victor Guilherme~Turrisi da~Costa, Louis Béthune, Zhe Gan, Alexander~T Toshev, Marcin Eichner, Moin Nabi, Yinfei Yang, Joshua~M. Susskind, and Alaaeldin El-Nouby.
\newblock Multimodal autoregressive pre-training of large vision encoders, 2024.
\newblock URL \url{https://arxiv.org/abs/2411.14402}.

\bibitem[Foster et~al.(2023)Foster, Croitoru, Dorfman, Edlund, Varsavsky, and Almazán]{foster2023flexible}
Thomas Foster, Ioana Croitoru, Robert Dorfman, Christoffer Edlund, Thomas Varsavsky, and Jon Almazán.
\newblock Flexible visual prompts for in-context learning in computer vision, 2023.

\bibitem[Ghiasi et~al.(2021)Ghiasi, Cui, Srinivas, Qian, Lin, Cubuk, Le, and Zoph]{ghiasi2021simplecopypastestrongdata}
Golnaz Ghiasi, Yin Cui, Aravind Srinivas, Rui Qian, Tsung-Yi Lin, Ekin~D. Cubuk, Quoc~V. Le, and Barret Zoph.
\newblock Simple copy-paste is a strong data augmentation method for instance segmentation, 2021.
\newblock URL \url{https://arxiv.org/abs/2012.07177}.

\bibitem[Goodfellow et~al.(2014)Goodfellow, Pouget-Abadie, Mirza, Xu, Warde-Farley, Ozair, Courville, and Bengio]{goodfellow2014generativeadversarialnetworks}
Ian~J. Goodfellow, Jean Pouget-Abadie, Mehdi Mirza, Bing Xu, David Warde-Farley, Sherjil Ozair, Aaron Courville, and Yoshua Bengio.
\newblock Generative adversarial networks, 2014.
\newblock URL \url{https://arxiv.org/abs/1406.2661}.

\bibitem[Goyal et~al.(2017)Goyal, Khot, Summers-Stay, Batra, and Parikh]{balanced_vqa_v2}
Yash Goyal, Tejas Khot, Douglas Summers-Stay, Dhruv Batra, and Devi Parikh.
\newblock Making the {V} in {VQA} matter: Elevating the role of image understanding in {V}isual {Q}uestion {A}nswering.
\newblock In \emph{Conference on Computer Vision and Pattern Recognition (CVPR)}, 2017.

\bibitem[Hahn et~al.(2025)Hahn, Reich, Araslanov, Cremers, Rupprecht, and Roth]{hahn2025scenecentricunsupervisedpanopticsegmentation}
Oliver Hahn, Christoph Reich, Nikita Araslanov, Daniel Cremers, Christian Rupprecht, and Stefan Roth.
\newblock Scene-centric unsupervised panoptic segmentation, 2025.
\newblock URL \url{https://arxiv.org/abs/2504.01955}.

\bibitem[He et~al.(2021)He, Chen, Xie, Li, Doll{\'a}r, and Girshick]{MAE}
Kaiming He, Xinlei Chen, Saining Xie, Yanghao Li, Piotr Doll{\'a}r, and Ross Girshick.
\newblock Masked autoencoders are scalable vision learners.
\newblock \emph{arXiv preprint arXiv:2111.06377}, 2021.

\bibitem[He et~al.(2022)He, Chen, Xie, Li, Doll{\'a}r, and Girshick]{he2022masked}
Kaiming He, Xinlei Chen, Saining Xie, Yanghao Li, Piotr Doll{\'a}r, and Ross Girshick.
\newblock Masked autoencoders are scalable vision learners.
\newblock In \emph{Proceedings of the IEEE/CVF conference on computer vision and pattern recognition}, pages 16000--16009, 2022.

\bibitem[Ho et~al.(2020)Ho, Jain, and Abbeel]{ho2020denoisingdiffusionprobabilisticmodels}
Jonathan Ho, Ajay Jain, and Pieter Abbeel.
\newblock Denoising diffusion probabilistic models, 2020.
\newblock URL \url{https://arxiv.org/abs/2006.11239}.

\bibitem[Hudson and Manning(2019)]{hudson2019gqanewdatasetrealworld}
Drew~A. Hudson and Christopher~D. Manning.
\newblock Gqa: A new dataset for real-world visual reasoning and compositional question answering, 2019.
\newblock URL \url{https://arxiv.org/abs/1902.09506}.

\bibitem[Kirillov et~al.(2023)Kirillov, Mintun, Ravi, Mao, Rolland, Gustafson, Xiao, Whitehead, Berg, Lo, et~al.]{kirillov2023segment}
Alexander Kirillov, Eric Mintun, Nikhila Ravi, Hanzi Mao, Chloe Rolland, Laura Gustafson, Tete Xiao, Spencer Whitehead, Alexander~C Berg, Wan-Yen Lo, et~al.
\newblock Segment anything.
\newblock \emph{arXiv preprint arXiv:2304.02643}, 2023.

\bibitem[Kojima et~al.(2023)Kojima, Gu, Reid, Matsuo, and Iwasawa]{kojima2023largelanguagemodelszeroshot}
Takeshi Kojima, Shixiang~Shane Gu, Machel Reid, Yutaka Matsuo, and Yusuke Iwasawa.
\newblock Large language models are zero-shot reasoners, 2023.
\newblock URL \url{https://arxiv.org/abs/2205.11916}.

\bibitem[Krizhevsky(2009)]{Krizhevsky2009LearningML}
Alex Krizhevsky.
\newblock Learning multiple layers of features from tiny images.
\newblock 2009.
\newblock URL \url{https://api.semanticscholar.org/CorpusID:18268744}.

\bibitem[Lai et~al.(2025)Lai, Zhong, Su, and Yang]{lai2025patientspecificautoregressivemodelsorgan}
Yuxiang Lai, Jike Zhong, Vanessa Su, and Xiaofeng Yang.
\newblock Patient-specific autoregressive models for organ motion prediction in radiotherapy, 2025.
\newblock URL \url{https://arxiv.org/abs/2505.11832}.

\bibitem[Li et~al.(2023{\natexlab{a}})Li, Jiang, Zhang, Ren, Liu, Zou, Xu, Li, Li, Yang, Zhang, and Gao]{li2023visual}
Feng Li, Qing Jiang, Hao Zhang, Tianhe Ren, Shilong Liu, Xueyan Zou, Huaizhe Xu, Hongyang Li, Chunyuan Li, Jianwei Yang, Lei Zhang, and Jianfeng Gao.
\newblock Visual in-context prompting, 2023{\natexlab{a}}.

\bibitem[Li et~al.(2022{\natexlab{a}})Li, Zheng, Liu, Wang, Su, and Zheng]{li2022semmae}
Gang Li, Heliang Zheng, Daqing Liu, Chaoyue Wang, Bing Su, and Changwen Zheng.
\newblock Semmae: Semantic-guided masking for learning masked autoencoders.
\newblock \emph{Advances in Neural Information Processing Systems}, 35:\penalty0 14290--14302, 2022{\natexlab{a}}.

\bibitem[Li et~al.(2022{\natexlab{b}})Li, Zheng, Liu, Wang, Su, and Zheng]{semae}
Gang Li, Heliang Zheng, Daqing Liu, Chaoyue Wang, Bing Su, and Changwen Zheng.
\newblock Semmae: Semantic-guided masking for learning masked autoencoders, 2022{\natexlab{b}}.

\bibitem[Li et~al.(2022{\natexlab{c}})Li, Li, Xiong, and Hoi]{li2022blip}
Junnan Li, Dongxu Li, Caiming Xiong, and Steven Hoi.
\newblock Blip: Bootstrapping language-image pre-training for unified vision-language understanding and generation, 2022{\natexlab{c}}.

\bibitem[Li et~al.(2023{\natexlab{b}})Li, Li, Savarese, and Hoi]{li2023blip2}
Junnan Li, Dongxu Li, Silvio Savarese, and Steven Hoi.
\newblock Blip-2: Bootstrapping language-image pre-training with frozen image encoders and large language models, 2023{\natexlab{b}}.

\bibitem[Li et~al.(2024)Li, Zhong, Li, Li, Lin, and Sugiyama]{li2024visionlanguagemodelfinetuningsimple}
Ming Li, Jike Zhong, Chenxin Li, Liuzhuozheng Li, Nie Lin, and Masashi Sugiyama.
\newblock Vision-language model fine-tuning via simple parameter-efficient modification, 2024.
\newblock URL \url{https://arxiv.org/abs/2409.16718}.

\bibitem[Li et~al.(2022{\natexlab{d}})Li, Ge, Yi, Hu, Shan, and Duan]{mc-BEiT}
Xiaotong Li, Yixiao Ge, Kun Yi, Zixuan Hu, Ying Shan, and Ling-Yu Duan.
\newblock mc-beit: Multi-choice discretization for image bert pre-training.
\newblock \emph{arXiv preprint arXiv:2203.15371}, 2022{\natexlab{d}}.

\bibitem[Liu et~al.(2023)Liu, Li, Wu, and Lee]{liu2023visualinstructiontuning}
Haotian Liu, Chunyuan Li, Qingyang Wu, and Yong~Jae Lee.
\newblock Visual instruction tuning, 2023.
\newblock URL \url{https://arxiv.org/abs/2304.08485}.

\bibitem[Liu et~al.(2024)Liu, Chen, Tian, Zou, Chen, and Cui]{liu2024llmconversationalagentmemory}
Na~Liu, Liangyu Chen, Xiaoyu Tian, Wei Zou, Kaijiang Chen, and Ming Cui.
\newblock From llm to conversational agent: A memory enhanced architecture with fine-tuning of large language models, 2024.
\newblock URL \url{https://arxiv.org/abs/2401.02777}.

\bibitem[Liu et~al.(2021)Liu, Lin, Cao, Hu, Wei, Zhang, Lin, and Guo]{liu2021swintransformerhierarchicalvision}
Ze~Liu, Yutong Lin, Yue Cao, Han Hu, Yixuan Wei, Zheng Zhang, Stephen Lin, and Baining Guo.
\newblock Swin transformer: Hierarchical vision transformer using shifted windows, 2021.
\newblock URL \url{https://arxiv.org/abs/2103.14030}.

\bibitem[Loshchilov and Hutter(2017{\natexlab{a}})]{AdamW}
Ilya Loshchilov and Frank Hutter.
\newblock Decoupled weight decay regularization.
\newblock \emph{arXiv preprint arXiv:1711.05101}, 2017{\natexlab{a}}.

\bibitem[Loshchilov and Hutter(2017{\natexlab{b}})]{loshchilov2017decoupled}
Ilya Loshchilov and Frank Hutter.
\newblock Decoupled weight decay regularization.
\newblock \emph{arXiv preprint arXiv:1711.05101}, 2017{\natexlab{b}}.

\bibitem[Lu et~al.(2022)Lu, Mishra, Xia, Qiu, Chang, Zhu, Tafjord, Clark, and Kalyan]{lu2022learnexplainmultimodalreasoning}
Pan Lu, Swaroop Mishra, Tony Xia, Liang Qiu, Kai-Wei Chang, Song-Chun Zhu, Oyvind Tafjord, Peter Clark, and Ashwin Kalyan.
\newblock Learn to explain: Multimodal reasoning via thought chains for science question answering, 2022.
\newblock URL \url{https://arxiv.org/abs/2209.09513}.

\bibitem[Ma et~al.(2022)Ma, Lu, Huang, Yang, Xu, Mo, Ren, and Li]{ma2022advanced}
Xiaoxiao Ma, Xinai Lu, Yihong Huang, Xinyi Yang, Ziyin Xu, Guozhao Mo, Yufei Ren, and Lin Li.
\newblock An advanced chicken face detection network based on gan and mae.
\newblock \emph{Animals}, 12\penalty0 (21):\penalty0 3055, 2022.

\bibitem[Martin(2007)]{Martin2007}
A.~Martin.
\newblock The representation of object concepts in the brain.
\newblock \emph{Annual Review of Psychology}, 58:\penalty0 25--45, 2007.

\bibitem[Naeem et~al.(2023)Naeem, Xian, Zhai, Hoyer, Gool, and Tombari]{naeem2023silcimprovingvisionlanguage}
Muhammad~Ferjad Naeem, Yongqin Xian, Xiaohua Zhai, Lukas Hoyer, Luc~Van Gool, and Federico Tombari.
\newblock Silc: Improving vision language pretraining with self-distillation, 2023.
\newblock URL \url{https://arxiv.org/abs/2310.13355}.

\bibitem[OpenAI(2023)]{openai2023chatgpt}
OpenAI.
\newblock Chatgpt: Gpt-3.5.
\newblock Online, 2023.
\newblock Available at \url{https://openai.com/chatgpt}.

\bibitem[OpenAI and team(2024)]{openai2024gpt4technicalreport}
OpenAI and GPT~4 team.
\newblock Gpt-4 technical report, 2024.
\newblock URL \url{https://arxiv.org/abs/2303.08774}.

\bibitem[Pathak et~al.(2016)Pathak, Krahenbuhl, Donahue, Darrell, and Efros]{pathak2016context}
Deepak Pathak, Philipp Krahenbuhl, Jeff Donahue, Trevor Darrell, and Alexei~A Efros.
\newblock Context encoders: Feature learning by inpainting.
\newblock In \emph{CVPR}, pages 2536--2544, 2016.

\bibitem[Radford et~al.(2021)Radford, Kim, Hallacy, Ramesh, Goh, Agarwal, Sastry, Askell, Mishkin, Clark, et~al.]{CLIP}
Alec Radford, Jong~Wook Kim, Chris Hallacy, Aditya Ramesh, Gabriel Goh, Sandhini Agarwal, Girish Sastry, Amanda Askell, Pamela Mishkin, Jack Clark, et~al.
\newblock Learning transferable visual models from natural language supervision.
\newblock In \emph{ICML}, pages 8748--8763, 2021.

\bibitem[Reppa et~al.(2020)Reppa, Williams, Greville, and Saunders]{reppa2020relative}
Irene Reppa, Kate~E Williams, W~James Greville, and Jo~Saunders.
\newblock The relative contribution of shape and colour to object memory.
\newblock \emph{Memory \& Cognition}, 48:\penalty0 1504--1521, 2020.

\bibitem[Schaeffer et~al.(2023)Schaeffer, Miranda, and Koyejo]{schaeffer2023emergentabilitieslargelanguage}
Rylan Schaeffer, Brando Miranda, and Sanmi Koyejo.
\newblock Are emergent abilities of large language models a mirage?, 2023.
\newblock URL \url{https://arxiv.org/abs/2304.15004}.

\bibitem[Shaban et~al.(2017)Shaban, Bansal, Liu, Essa, and Boots]{shaban2017one}
Amirreza Shaban, Shray Bansal, Zhen Liu, Irfan Essa, and Byron Boots.
\newblock One-shot learning for semantic segmentation.
\newblock \emph{arXiv preprint arXiv:1709.03410}, 2017.

\bibitem[Sheng et~al.(2023)Sheng, Chen, Tan, Liu, Chu, Bao, Gong, Liu, Xu, and Yu]{sheng2023unified}
Dianmo Sheng, Dongdong Chen, Zhentao Tan, Qiankun Liu, Qi~Chu, Jianmin Bao, Tao Gong, Bin Liu, Shengwei Xu, and Nenghai Yu.
\newblock Towards more unified in-context visual understanding, 2023.

\bibitem[Sun et~al.(2023)Sun, Chen, Wang, Wang, and Li]{sun2023exploring}
Yanpeng Sun, Qiang Chen, Jian Wang, Jingdong Wang, and Zechao Li.
\newblock Exploring effective factors for improving visual in-context learning, 2023.

\bibitem[Tao et~al.(2022)Tao, Zhu, Su, Huang, Li, Zhou, Qiao, Wang, and Dai]{tao2022siamese}
Chenxin Tao, Xizhou Zhu, Weijie Su, Gao Huang, Bin Li, Jie Zhou, Yu~Qiao, Xiaogang Wang, and Jifeng Dai.
\newblock Siamese image modeling for self-supervised vision representation learning, 2022.

\bibitem[Team(2024{\natexlab{a}})]{geminiteam2024gemini15unlockingmultimodal}
Gemini Team.
\newblock Gemini 1.5: Unlocking multimodal understanding across millions of tokens of context, 2024{\natexlab{a}}.
\newblock URL \url{https://arxiv.org/abs/2403.05530}.

\bibitem[Team(2024{\natexlab{b}})]{geminiteam2024geminifamilyhighlycapable}
Gemini Team.
\newblock Gemini: A family of highly capable multimodal models, 2024{\natexlab{b}}.
\newblock URL \url{https://arxiv.org/abs/2312.11805}.

\bibitem[Tong et~al.(2024)Tong, Brown, Wu, Woo, Middepogu, Akula, Yang, Yang, Iyer, Pan, Wang, Fergus, LeCun, and Xie]{tong2024cambrian1fullyopenvisioncentric}
Shengbang Tong, Ellis Brown, Penghao Wu, Sanghyun Woo, Manoj Middepogu, Sai~Charitha Akula, Jihan Yang, Shusheng Yang, Adithya Iyer, Xichen Pan, Ziteng Wang, Rob Fergus, Yann LeCun, and Saining Xie.
\newblock Cambrian-1: A fully open, vision-centric exploration of multimodal llms, 2024.
\newblock URL \url{https://arxiv.org/abs/2406.16860}.

\bibitem[Tukra et~al.(2023)Tukra, Hoffman, and Chatfield]{tukra2023improving}
Samyakh Tukra, Frederick Hoffman, and Ken Chatfield.
\newblock Improving visual representation learning through perceptual understanding.
\newblock In \emph{Proceedings of the IEEE/CVF Conference on Computer Vision and Pattern Recognition}, pages 14486--14495, 2023.

\bibitem[Vaswani et~al.(2017)Vaswani, Shazeer, Parmar, Uszkoreit, Jones, Gomez, Kaiser, and Polosukhin]{vaswani2017attention}
Ashish Vaswani, Noam Shazeer, Niki Parmar, Jakob Uszkoreit, Llion Jones, Aidan~N Gomez, {\L}ukasz Kaiser, and Illia Polosukhin.
\newblock Attention is all you need.
\newblock \emph{Advances in neural information processing systems}, 30, 2017.

\bibitem[Vincent et~al.(2008)Vincent, Larochelle, Bengio, and Manzagol]{Vincent2008}
P.~Vincent, H.~Larochelle, Y.~Bengio, and P.-A. Manzagol.
\newblock Extracting and composing robust features with denoising autoencoders.
\newblock In \emph{International Conference on Machine Learning proceedings}. 2008.

\bibitem[Wan et~al.(2024)Wan, Tschannen, Xian, Pavetic, Alabdulmohsin, Wang, Pinto, Steiner, Beyer, and Zhai]{wan2024loccavisualpretraininglocationaware}
Bo~Wan, Michael Tschannen, Yongqin Xian, Filip Pavetic, Ibrahim Alabdulmohsin, Xiao Wang, André~Susano Pinto, Andreas Steiner, Lucas Beyer, and Xiaohua Zhai.
\newblock Locca: Visual pretraining with location-aware captioners, 2024.
\newblock URL \url{https://arxiv.org/abs/2403.19596}.

\bibitem[Wang et~al.(2024)Wang, Ma, Feng, Zhang, Yang, Zhang, Chen, Tang, Chen, Lin, Zhao, Wei, and Wen]{Wang_2024}
Lei Wang, Chen Ma, Xueyang Feng, Zeyu Zhang, Hao Yang, Jingsen Zhang, Zhiyuan Chen, Jiakai Tang, Xu~Chen, Yankai Lin, Wayne~Xin Zhao, Zhewei Wei, and Jirong Wen.
\newblock A survey on large language model based autonomous agents.
\newblock \emph{Frontiers of Computer Science}, 18\penalty0 (6), March 2024.
\newblock ISSN 2095-2236.
\newblock \doi{10.1007/s11704-024-40231-1}.
\newblock URL \url{http://dx.doi.org/10.1007/s11704-024-40231-1}.

\bibitem[Wang et~al.(2021)Wang, Xie, Li, Fan, Song, Liang, Lu, Luo, and Shao]{wang2021pyramidvisiontransformerversatile}
Wenhai Wang, Enze Xie, Xiang Li, Deng-Ping Fan, Kaitao Song, Ding Liang, Tong Lu, Ping Luo, and Ling Shao.
\newblock Pyramid vision transformer: A versatile backbone for dense prediction without convolutions, 2021.
\newblock URL \url{https://arxiv.org/abs/2102.12122}.

\bibitem[Wang et~al.(2023{\natexlab{a}})Wang, Wang, Cao, Shen, and Huang]{painter}
Xinlong Wang, Wen Wang, Yue Cao, Chunhua Shen, and Tiejun Huang.
\newblock Images speak in images: A generalist painter for in-context visual learning, 2023{\natexlab{a}}.

\bibitem[Wang et~al.(2023{\natexlab{b}})Wang, Zhang, Cao, Wang, Shen, and Huang]{wang2023seggpt}
Xinlong Wang, Xiaosong Zhang, Yue Cao, Wen Wang, Chunhua Shen, and Tiejun Huang.
\newblock Seggpt: Segmenting everything in context, 2023{\natexlab{b}}.

\bibitem[Wang and Zhou(2024)]{wang2024chainofthoughtreasoningprompting}
Xuezhi Wang and Denny Zhou.
\newblock Chain-of-thought reasoning without prompting, 2024.
\newblock URL \url{https://arxiv.org/abs/2402.10200}.

\bibitem[Wei et~al.(2022)Wei, Tay, Bommasani, Raffel, Zoph, Borgeaud, Yogatama, Bosma, Zhou, Metzler, Chi, Hashimoto, Vinyals, Liang, Dean, and Fedus]{wei2022emergentabilitieslargelanguage}
Jason Wei, Yi~Tay, Rishi Bommasani, Colin Raffel, Barret Zoph, Sebastian Borgeaud, Dani Yogatama, Maarten Bosma, Denny Zhou, Donald Metzler, Ed~H. Chi, Tatsunori Hashimoto, Oriol Vinyals, Percy Liang, Jeff Dean, and William Fedus.
\newblock Emergent abilities of large language models, 2022.
\newblock URL \url{https://arxiv.org/abs/2206.07682}.

\bibitem[Wei et~al.(2023)Wei, Wang, Schuurmans, Bosma, Ichter, Xia, Chi, Le, and Zhou]{wei2023chainofthoughtpromptingelicitsreasoning}
Jason Wei, Xuezhi Wang, Dale Schuurmans, Maarten Bosma, Brian Ichter, Fei Xia, Ed~Chi, Quoc Le, and Denny Zhou.
\newblock Chain-of-thought prompting elicits reasoning in large language models, 2023.
\newblock URL \url{https://arxiv.org/abs/2201.11903}.

\bibitem[Wiedemer et~al.(2025)Wiedemer, Li, Vicol, Gu, Matarese, Swersky, Kim, Jaini, and Geirhos]{wiedemer2025videomodelszeroshotlearners}
Thaddäus Wiedemer, Yuxuan Li, Paul Vicol, Shixiang~Shane Gu, Nick Matarese, Kevin Swersky, Been Kim, Priyank Jaini, and Robert Geirhos.
\newblock Video models are zero-shot learners and reasoners, 2025.
\newblock URL \url{https://arxiv.org/abs/2509.20328}.

\bibitem[Wu et~al.(2021)Wu, Xiao, Codella, Liu, Dai, Yuan, and Zhang]{wu2021cvtintroducingconvolutionsvision}
Haiping Wu, Bin Xiao, Noel Codella, Mengchen Liu, Xiyang Dai, Lu~Yuan, and Lei Zhang.
\newblock Cvt: Introducing convolutions to vision transformers, 2021.
\newblock URL \url{https://arxiv.org/abs/2103.15808}.

\bibitem[Yang et~al.(2021)Yang, Li, Zhang, Dai, Xiao, Yuan, and Gao]{yang2021focalselfattentionlocalglobalinteractions}
Jianwei Yang, Chunyuan Li, Pengchuan Zhang, Xiyang Dai, Bin Xiao, Lu~Yuan, and Jianfeng Gao.
\newblock Focal self-attention for local-global interactions in vision transformers, 2021.
\newblock URL \url{https://arxiv.org/abs/2107.00641}.

\bibitem[Zhang et~al.(2022)Zhang, Pan, Chen, Zhong, Fu, and Chao]{zhang2022learningfreeobjectsegments}
Cheng Zhang, Tai-Yu Pan, Tianle Chen, Jike Zhong, Wenjin Fu, and Wei-Lun Chao.
\newblock Learning with free object segments for long-tailed instance segmentation, 2022.
\newblock URL \url{https://arxiv.org/abs/2202.11124}.

\bibitem[Zhang et~al.(2023{\natexlab{a}})Zhang, Wang, Li, Nakashima, and Nagahara]{zhang2023instruct}
Jiahao Zhang, Bowen Wang, Liangzhi Li, Yuta Nakashima, and Hajime Nagahara.
\newblock Instruct me more! random prompting for visual in-context learning, 2023{\natexlab{a}}.

\bibitem[Zhang et~al.(2023{\natexlab{b}})Zhang, Zhou, and Liu]{zhang2023makes}
Yuanhan Zhang, Kaiyang Zhou, and Ziwei Liu.
\newblock What makes good examples for visual in-context learning?, 2023{\natexlab{b}}.

\bibitem[Zhao et~al.(2024)Zhao, Zhou, Li, Tang, Wang, Hou, Min, Zhang, Zhang, Dong, Du, Yang, Chen, Chen, Jiang, Ren, Li, Tang, Liu, Liu, Nie, and Wen]{zhao2024surveylargelanguagemodels}
Wayne~Xin Zhao, Kun Zhou, Junyi Li, Tianyi Tang, Xiaolei Wang, Yupeng Hou, Yingqian Min, Beichen Zhang, Junjie Zhang, Zican Dong, Yifan Du, Chen Yang, Yushuo Chen, Zhipeng Chen, Jinhao Jiang, Ruiyang Ren, Yifan Li, Xinyu Tang, Zikang Liu, Peiyu Liu, Jian-Yun Nie, and Ji-Rong Wen.
\newblock A survey of large language models, 2024.
\newblock URL \url{https://arxiv.org/abs/2303.18223}.

\bibitem[Zhou et~al.(2021)Zhou, Wei, Wang, Shen, Xie, Yuille, and Kong]{iBoT}
Jinghao Zhou, Chen Wei, Huiyu Wang, Wei Shen, Cihang Xie, Alan Yuille, and Tao Kong.
\newblock ibot: Image bert pre-training with online tokenizer.
\newblock \emph{arXiv preprint arXiv:2111.07832}, 2021.

\end{thebibliography}

\clearpage
\appendix

\section*{Appendix}

\label{supplementary}
We provide all additional details for our paper in the following sections. 

\begin{itemize}
    \item Border Impact. We discuss the limitations and potential future follow-up work. 
    \item Details of the Implementation. We provide additional details of model setup, training schedules. 
    \item Ablation Studies. We provide additional ablation study results, including masking strategies, model size, and object-mask ratio.
    \item Discussions. We address additional questions about the usage of additional data, the generalization capability of our proposed tokenization objective, as well as impact of auxiliary Gan loss. 
\end{itemize}

\section{Broader Impact}
\label{Impact}

\begin{figure}[t]
    \centering
    \includegraphics[width=\linewidth]{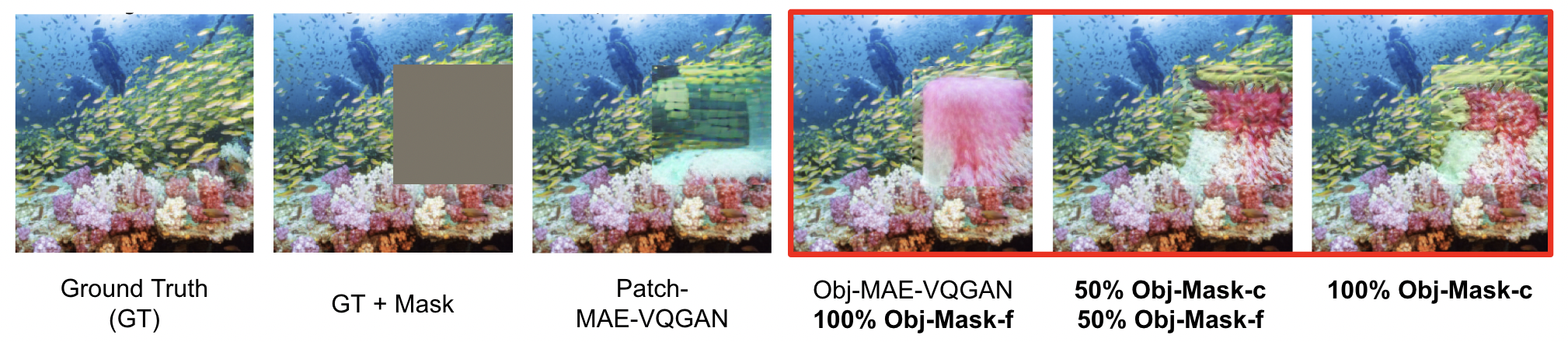}
   
    \caption{\textbf{Failure Cases:} (4): Failure case of reconstruction with fine-grained object masking (Obj-Mask-f). (5)-(6): Remedy by using coarse object masking (Obj-Mask-c)}
    \label{fig:faliure}
\end{figure}

\smallskip\noindent\textbf{Limitations and future work.}
While our method improves semantic reasoning, there are still some failure cases (\autoref{fig:faliure}). For example, when using fine-grained object masking during pre-training—where the mask follows the exact shape of objects—the model may "cheat" by overfitting to the mask shape. In such cases, it quickly learns to fill in the masked area without acquiring meaningful representations. To resolve this issue, we expand the mask to the bounding box. In future work, we aim to develop a more structured and robust tokenizer to enhance the model’s reasoning capabilities. In addition, we acknowledge the cost of segmentation overhead, but in our respectful opinion, our pipeline should be viewed as a proof-of-concept, and the performance gain is strong enough to justify studying it.

\smallskip\noindent\textbf{Ethics Statement.}
We ensure that our approach adheres to all legal and ethical guidelines throughout its development, with no violations. Fair compensation was provided to all annotators and graduate students involved in this work. The problems used in our study were collected from publicly accessible exams\footnote{https://gate2025.iitr.ac.in/} and resources licensed under CC Licenses\footnote{https://www.allaboutcircuits.com/worksheets/}\footnote{https://ocw.mit.edu/}. This research is conducted solely for academic purposes, and we strictly prohibit any commercial use of the results. Additionally, the spurious captions generated in Section 4 are limited to problem-solving contexts and pose no harm to individuals.

\smallskip\noindent\textbf{Reproducibility statement.} We are committed to efficient and reproducible research. All code, datasets, and models will be publicly released.

\section{Additional Implementation Details}
\label{Detail}
\smallskip\noindent\textbf{Mask generation and preprocessing.} To efficiently generate object masks, we leverage off-the-shelf \citep{kirillov2023segment}, a popular unsupervised segmentation model, to infer scene-centric images (where many objects are present). This step yields a set of binary object masks, which we then convert into the COCO RLE (Run-Length Encoding) format. Note that this step can be done either online (during the forward pass of each batch) or beforehand. Here we test both and empirically find the pre-processing step crucial as it saves $3\times$ GPU hours as shown in \autoref{tab:cost}. This solution is scalable as more data can be generated directly using the pre-trained SAM model.

\begin{table}[h]
    
    \centering
    \small
    \begin{tabular}{c|c|c}
    \toprule
        Model & Pre-Processing & Training Cost \\
    \midrule
        MIM (w. Obj Rep) & \checkmark & 3.6 {\textcolor{redxx}{($-2.7\times$)}}\\
        MIM (w. Obj Rep) & $\times$ & 9.8 \\
    \midrule
        MIM+VQGAN(w. Obj Rep) & \checkmark &  5.1 {\textcolor{redxx}{($-2.5\times$)}}\\
        MIM+VQGAN(w. Obj Rep) & $\times$ & 13.2 \\
    \bottomrule
    \end{tabular}
    \caption{Comparison of training costs in GPU hours with and without pre-processing for 1 epoch training using $500K$ data and a single A100 GPU.}
    \label{tab:cost}
\end{table}

\smallskip\noindent\textbf{Implementation details on downstream tasks.} Following \citet{he2022masked}, we first discard the decoder after pre-training is complete. For end-to-end FT, we use AdamW \citep{loshchilov2017decoupled} optimizer with base learning rate $blr=1.0\times10^{-3}$, weight decay $0.05$, layer decay $0.75$ and train for $20$ epochs with $5$ rounds of warmup epochs. Additionally, we use drop path $0.1$ with mixup $0.8$ and ensure the effective batch size is $1024$ by accumulating SGD iters. For LP, we use base learning rate $blr=1.0\times10^{-1}$ and an effective batch size of $16384$ while keeping other settings the same. In our model. each self-attention layer includes $\alpha = 16$ attention heads.

\smallskip\noindent\textbf{Implementation details on pertaining.} For the first stage, we use AdamW \citep{loshchilov2017decoupled} optimizer with a base learning of $blr=1.5\times10^{-4}$, weight decay $wd=0.05$, and the cosine learning rate decay scheduler. We accumulate iterations to emulate the recommended batch size of 4096 and pre-train the model for 25 epochs with 5 warmup epochs. During this stage, the mask ratio is set for $mr_{patch}=75\%$. For the second stage, we start from the saved checkpoint from stage one. We apply an object ratio of $mr_\mathrm{obj}=50\%$ which randomly masks out $25$ objects in each image by hiding the patches spatially covering them. To enable batch processing, we apply an additional mask ratio constraint of $mr_{patch}=60\%$ on all images. The mask ratio is set $15\%$ lower to accommodate increased difficulty in the objective. 

Due to constraints in computing resources, we use publicly available pre-trained checkpoints\footnote{https://github.com/facebookresearch/mae}\footnote{https://github.com/amirbar/visual\_prompting} as the starting model for both stages of pre-training, unless otherwise specified. Importantly, using pre-trained checkpoints does not undermine our objective, as they are trained with a patch-level objective, which aligns with the first stage of our framework for learning low-level representations (Two Stage Learning Section). Essentially, we retrain these models on a different dataset with some adaptations.

\smallskip\noindent\textbf{Loss function for MIM-VQGAN.}  
MIM-VQGAN was proposed by \citet{bar2022visual} to study the effectiveness of visual prompting,  
which effectively shifted the MIM evaluation paradigm from fine-tuning on downstream tasks to direct output generation via prompting.  
This can be seen as a unified framework for vision tasks.  
Unlike \citet{MAE}, which computes the MSE loss by directly regressing on pixel values,  
MIM-VQGAN instead computes the cross-entropy (CE) loss on the corresponding patch value in the quantized codebook.  
This design effectively alleviates ambiguity in generation, as the codebook is discrete, unlike pixel values.  
Notably, the underlying objective—masked autoencoding—remains unchanged.  
Hence, MIM-VQGAN provides an effective way to directly compare our proposed method.  
In our experiments, we follow the implementation of \citet{bar2022visual}.

\section{Additional Ablation Study.} 
\label{ablation}

\begin{figure}[H]
    \centering
    \includegraphics[width=\linewidth]{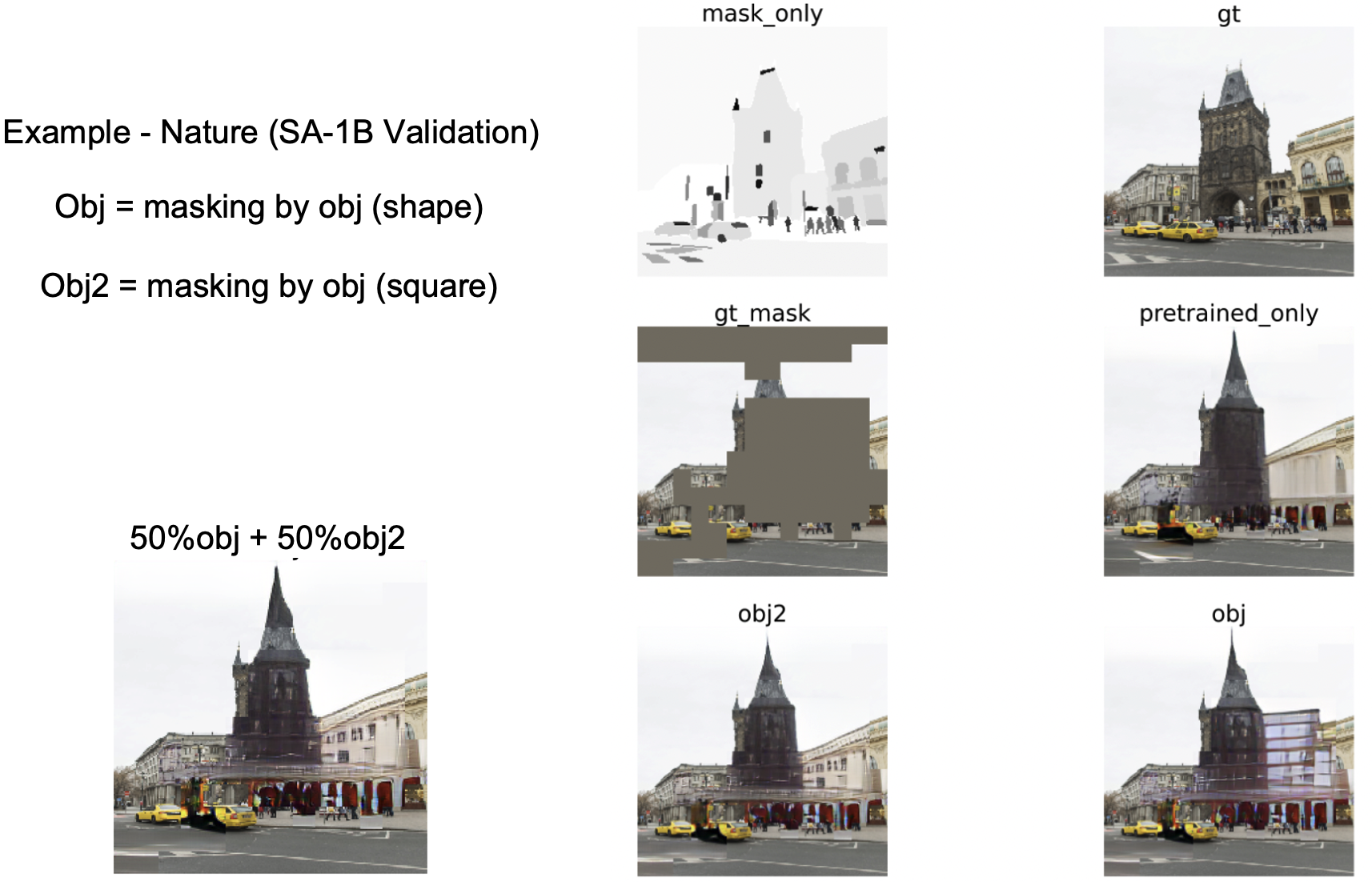}
   
    \caption{\textbf{Ablation Study of Masking Strategies (A)} }
    \label{fig:shape_square_1}
\end{figure}

\begin{figure}[t]
    \centering
    \includegraphics[width=\linewidth]{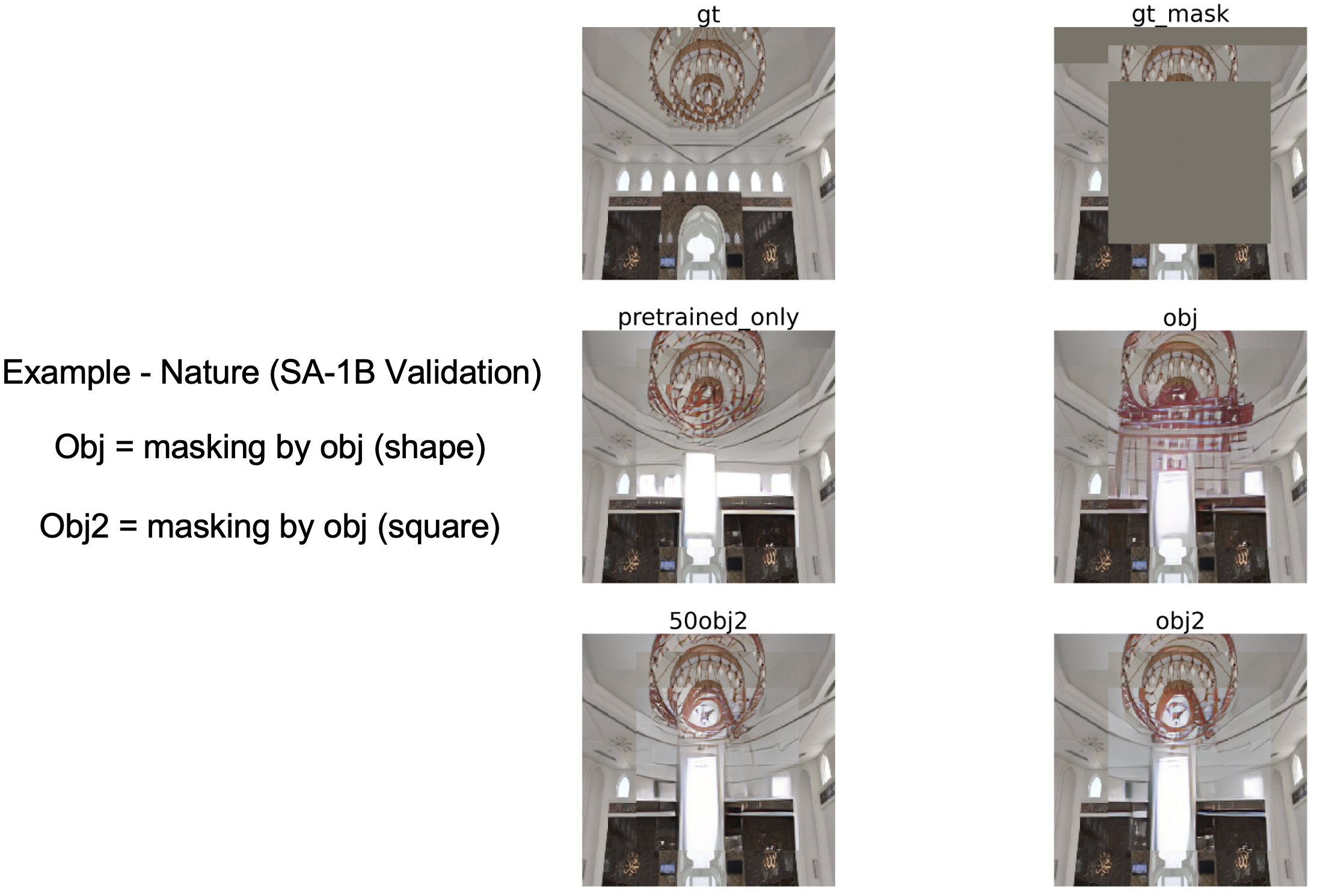}
   
    \caption{\textbf{Ablation Study of Masking Strategies (B)} }
    \label{fig:shape_square_2}
\end{figure}

\smallskip\noindent\textbf{Influence of different object masking strategies:}
As shown in~\autoref{fig:shape_square_1} and~\autoref{fig:shape_square_2}, we evaluate reconstruction performance using three masking strategies: masking strictly based on the object shape, masking the square region of the object, and a combination of both. While these visualizations demonstrate the superiority of object-based masking compared to random masking strategies, they also reveal certain limitations. Specifically, relying solely on object shape masking can lead to the model overfitting to the mask shape (“cheating”), while using only square masking results in sub-optimal performance on details. By combining these two strategies, we achieve more realistic and effective reconstruction.
\begin{figure}[t]
   
    \centering
    \includegraphics[width=0.7\linewidth]{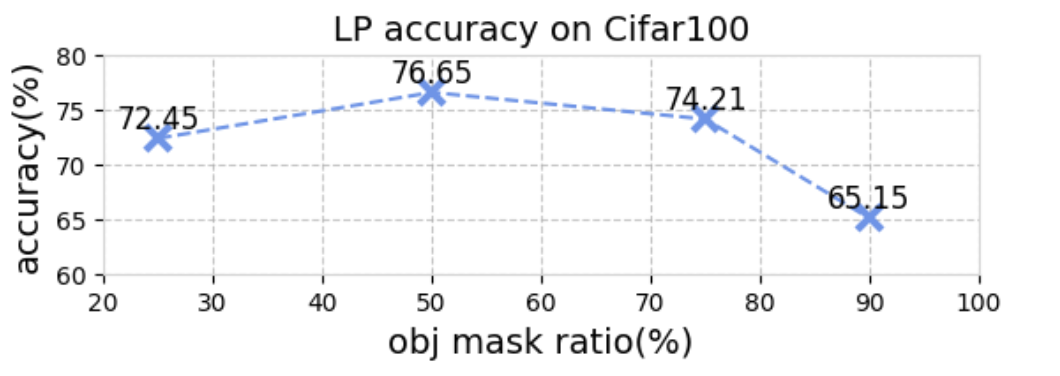}
    \caption{\textbf{Effect of object mask ratio:} The number of objects masked out during masked image modeling.
    }
    \label{fig:obj_ratio}
\end{figure}
\begin{table}[t]
    \centering
    \small
    \newcolumntype{C}[1]{>{\centering\arraybackslash}p{#1}}
    \begin{tabular}{C{2.5cm}|C{1.2cm}|C{1.5cm}C{1.5cm}}    
    \toprule
        \multirow{2}{*}{Model} & \multirow{2}{*}{Backbone} & \multicolumn{2}{c}{Cifar100 Top-1 Acc (\%)} \\
    \cmidrule{3-4}
          & & FT & LP \\
    \midrule
        $\text{MIM}^\dagger$ & ViT-B & 89.98 & 75.01 \\
        $\text{MIM}^\dagger$ & ViT-L & 92.67 & 76.20 \\
         $\text{MIM (w. Obj Rep)}$ & ViT-B & 90.08 & 72.44\\
         $\text{MIM (w. Obj Rep)} $ & ViT-L & 93.77 & 76.65\\
    \bottomrule
    \end{tabular}
    \caption{Comparison of different model sizes. Results show our approach is able to scale with model size.}
    \label{tab:vit size}
\end{table}
\begin{figure}[t]
    \centering
    \includegraphics[width=\linewidth]{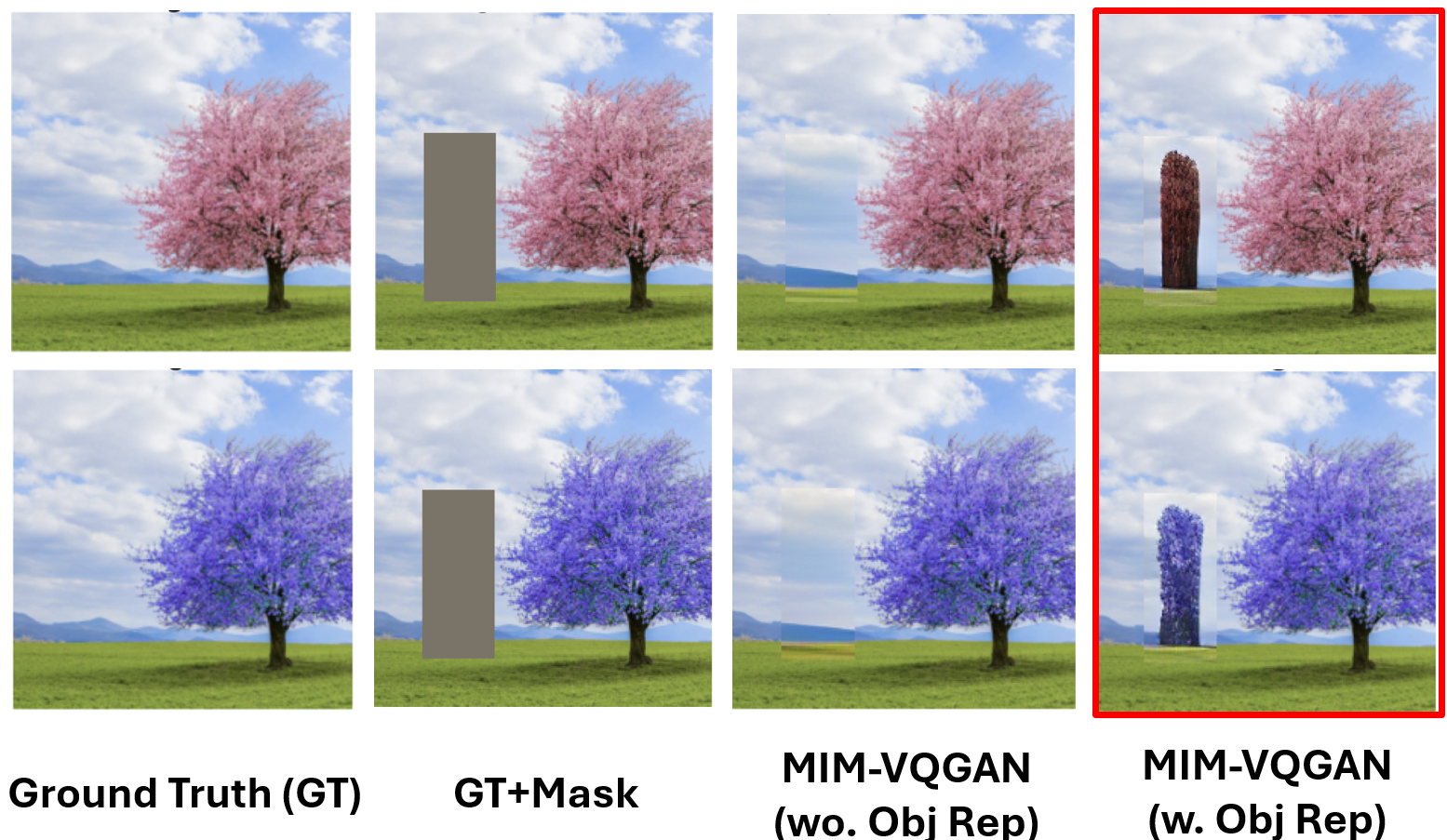}
   
    \caption{\textbf{Extend of color learning example}}
    \label{fig:object_color}
\end{figure}

\smallskip\noindent\textbf{Study on how the model captures context:} We investigate and visualize if our model has learned to capture the context during the pretraining process. Here we focus on learning the ``shape" and ``color", two of the most important ingredients to human learning. As we have addressed learning the ``shape" in \autoref{fig:color_shape_minimal} and Discussion Section, we showcase the learning of color in \autoref{fig:object_color}. In this example, when the same pair of examples but with different colors is given to the model, it is able to reconstruct objects of colors similar to the example, meaning that it does not infer color based on memorization but rather from the context that is given.

\smallskip\noindent\textbf{Study on model sizes:} \autoref{tab:vit size} shows the LP and FT results on different vit base models, and the result shows our observations and findings in Quantitative Evaluation and Discussion sections hold for different model sizes.

\smallskip\noindent\textbf{Obj-Mask Ratio.} 
To determine the influence of the masking strategy, we train our model with different mask ratios, as shown in \autoref{fig:obj_ratio}. Unlike traditional random patch-level masking, as in \citet{he2022masked}, object-level masking becomes less effective when obj-mask ratios exceed 50\%. This decline occurs because random masking often leaves portions of objects visible, which can help guide reconstruction, while object-level masking requires the model to learn the semantic relationships between objects only from other objects. We note that a 50\% obj-mask ratio effectively masks out around 75\% of the image.

\smallskip\noindent\textbf{Loss functions.} We further ablate the effect of object balance loss defined in \autoref{eq:allloss}. Results in \autoref{tab:loss_ablation} shows that combining both $\sL_{MIM}$ and $\sL_{\mathrm{obj}}$ achieves the best performance. 
\begin{table}[H]
    \centering
    \small
    \begin{tabular}{l|c}
        \toprule
        Model Variant & VQA (v2.0) Acc. (\%) \\
        \midrule
        MIM (w. Obj Rep) & 53.02 \\
        + $\sL_{MIM}$ only & 55.44 \\
        + $\sL_{\mathrm{obj}}$ only & 52.48 \\
        + $\sL_{MIM}$ + $\sL_{\mathrm{obj}}$ (Eq.~\ref{eq:allloss}) & \textbf{56.89} \\
        \bottomrule
    \end{tabular}
    \caption{Effect of adding different loss terms in Eq.~\ref{eq:allloss} on VQA (v2.0). Combining both $\sL_{MIM}$ and $\sL_{\mathrm{obj}}$ achieves the best performance.}
    \label{tab:loss_ablation}
    \vskip -10pt
\end{table}

\section{Additional Discussions.}
\label{add_discussion} 

\smallskip\noindent\textbf{Model size.}  Here we show LP results on Cifar-100 classification with ViT-B and ViT-L. \autoref{tab:vit size} indicates that our approach is scalable with respect to increasing model sizes.

\smallskip\noindent\textbf{Additional motivation for using object-level representation.} Besides computer vision research, neuroscience studies have also found that the human brain uses an object-centric approach for visual recognition \citep{BartnikGroen2023, BonnerEpstein2021, Martin2007}. Within computer vision research, object segmentations have also been found to be helpful for tasks such as instance segmentation \citep{ghiasi2021simplecopypastestrongdata} and weakly supervised learning \citep{zhang2022learningfreeobjectsegments}. Hence, we conjecture “object” as a plausible candidate and explore it as the masking unit in MAE by simply masking out random objects and inpainting them instead of random patches.

\smallskip\noindent\textbf{Generalizability of object-centric objective.} The surprising result is that while Patch-MAE severely degrades downstream fine-tuning performance, Obj-MIM can recover such gap in a short GPU-hour, demonstrating that object-centric learning objective enables the learning of highly semantic and generalizable features where the original Patch-MIM cannot, especially given the underlying semantic difference (domain gap) between the datasets.

\begin{figure}[t]
    \centering
    \includegraphics[width=0.6\linewidth]{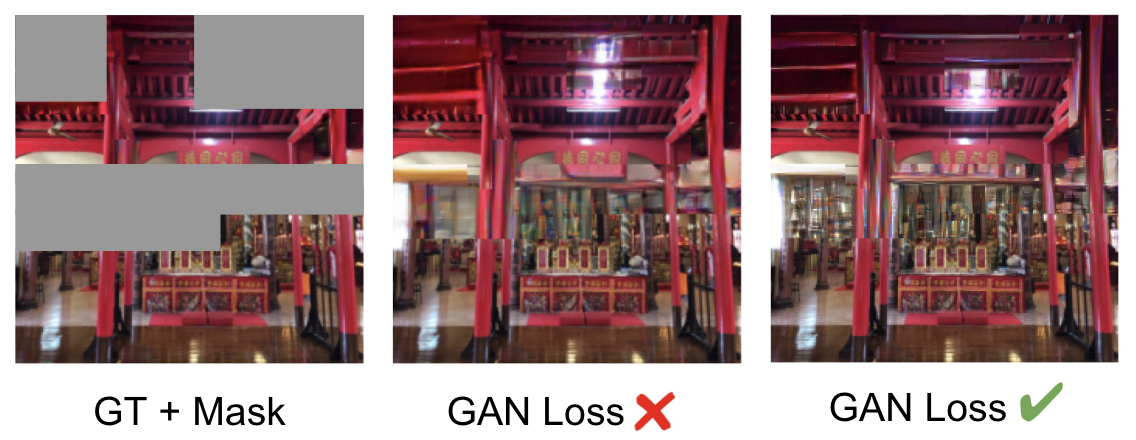}
   
    \caption{\textbf{GAN loss} can further help with better details.}
    \label{fig:gan loss}
\end{figure}
\smallskip\noindent\textbf{Furthur enhancing visual details with Gan loss.} Generative adversarial networks (GAN) 
\citep{goodfellow2014generativeadversarialnetworks} learn representation through the competition of a generator and a discriminator. Recent studies show that adding GAN losses can enhance visual details \citep{he2022masked, tukra2023improving, fei2023masked, ma2022advanced}. Following this intuition, we add an auxiliary GAN loss to our objective in \autoref{eq:allloss}:
\begin{equation}
\label{eq:alllosswgan}
\sL_{OBJ-MAE} = \sL_{MAE} + \lambda_{1}\cdot\sL_\mathrm{obj} + \lambda_{2}\cdot\sL_{GAN}
\end{equation}
This can be achieved by adding a simple discriminator and using the original network as the generator; details can be found in the Appendix. Results in (\autoref{fig:gan loss}) confirm that GAN loss can help produce more detailed images.
\end{document}